%% file: ms.tex
\documentclass[a4paper, conference]{IEEEtran}
\usepackage{cite}
\usepackage{amssymb}
\usepackage{url}
\usepackage[shortcuts,acronym]{glossaries}
\usepackage{hyperref}
\usepackage{tikz}
\usepackage{pgfplots}
\usepackage{float}
\usepackage[export]{adjustbox}
\usepackage[font=footnotesize]{subcaption}
\usetikzlibrary{matrix,shapes,arrows,positioning,chains,positioning}
\pgfplotsset{compat=1.14}

\makeglossaries

\newacronym{mav}{MAV}{Micro Aerial Vehicle}
\newacronym{sar}{SaR}{Search and Rescue}
\newacronym{pwm}{PWM}{Pulse Width Modulation}
\newacronym{ppm}{PPM}{Pulse Position Modulation}
\newacronym{gnss}{GNSS}{Global Navigation Satellite System}
\newacronym{gpio}{GPIO}{General Purpose Input/Output}
\newacronym{uart}{UART}{Universal Asynchronous Receiver/Transmitter}
\newacronym{lidar}{LIDAR}{LIght Detection And Ranging}

\begin{document}
\title{End to end collision avoidance based on optical flow and neural networks}

\author{\IEEEauthorblockN{Jan Blumenkamp}
\IEEEauthorblockA{University of Bremen}
}

\maketitle

\begin{abstract}
Optical flow is believed to play an important role in the agile flight of birds and insects. Even though it is a very simple concept, it is rarely used in computer vision for collision avoidance. This work implements a neural network based collision avoidance which was deployed and evaluated on a solely for this purpose refitted car. \textit{(6819 words)}
\end{abstract}

\IEEEpeerreviewmaketitle

\section{Introduction}
\label{sec:introduction}
Birds and insects are capable to fly with fast speeds and high accuracies with a simple vision system. It is believed that a core component in this agile flight is the evaluation of the optical flow \cite{dak_2016}. Even though this is a seemingly simple concept, current research is far from \glspl{mav} flying as agile as birds.

\begin{figure}[h]
  \centering
  \includegraphics[width=0.45\textwidth]{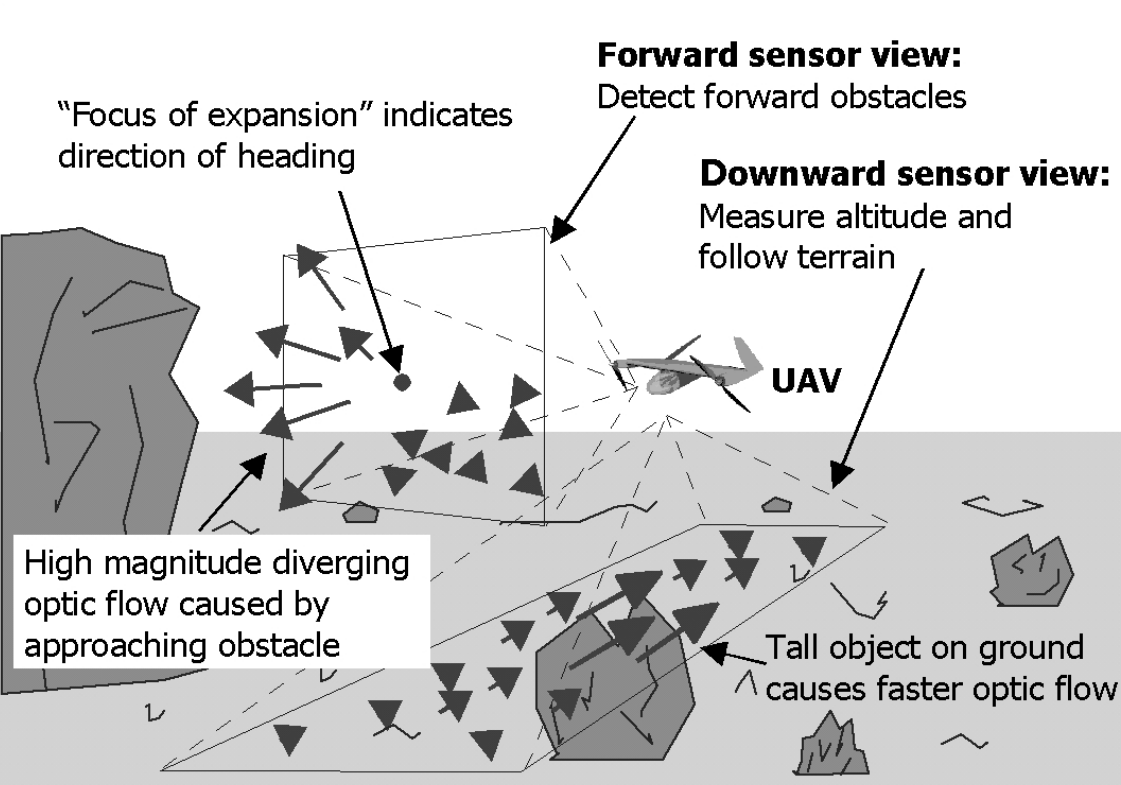}
  \caption{Visualization of the optical flow perceived by a flying robot: Different kind of optical flow from different perspectives. \cite{bar_2002}}
  \label{fig:opt_flow}
\end{figure}

The optical flow describes the movement of brightness changes of single pixels or areas of pixels in a frame, which is caused by the relative motion of the image sensor and viewed objects \cite{ber_1981}. This information can be used to not only detect motion, but also to reconstruct information of the three-dimensional structure of a scene, even with just a single camera. While it is impossible to reconstruct absolute distances and relations with this method, it is possible to re-interpret this information into a three-dimensional representation \cite{lon_1980}. Due to the fixed size of single objects, a frame can be segmented into objects moving in different relative speeds \cite{ber_1981}. The frontal and lateral optical flow is visualized in Fig. \ref{fig:opt_flow}.

Insects use optical flow clues for navigating through narrow passages by balancing the optical flow perceived on the left and right side, for controlling flight speed, altitude and attitude as well as for estimating distances. Among other birds, Hummingbirds may use optical flow to avoid collisions by monitoring obstacle attributes like verticals size, expansion and relative position. \cite{dak_2016}

The idea of using optical flow for collision avoidance is not entirely new. Previous research used this approach primarily in \glspl{mav}. \glspl{mav} have a broad set of use cases, from \glspl{sar} to delivery services. In these settings, \glspl{mav} need to be capable of flying close to the ground, through destroyed areas, in tunnels or caves or in an urban setting. \glspl{gnss} might not be available at all times, so an alternative solution has to be provided. Additionally, weight, size and speed plays an important role, requiring available sensors not only to be small, but also to be fast. \cite{bar_2002}

One of the first works that implemented a method for collision avoidance and landing in indoor environments based on optical flow utilized small optic flow microsensors \cite{gre_2003}. Later, \glspl{mav} were used to demonstrate the feasibility to fly outdoors and successfully avoid collisions with trees and other objects \cite{bey_2009}. Instead of a single microsensor, an array of optical mouse sensor with a custom designed optic was used to detect the optical flow in seven distinct regions. Embedded stereo camera can be used to estimate velocity and depth in combination with a distance sensor \cite{mcg_2017}. In one of the most recent works, \cite{san_2018} demonstrate the possibility to fly through narrow gaps by utilizing the optical flow.

All of these approaches have in common that they are analytical and focus on a specific use case, like frontal collision avoidance by evaluating the flow field divergence or following a corridor. With more powerful computer, machine learning has become more popular in the last years. \cite{mic_2005} uses a reinforcement learning approach to map measured distances to a camera image and based on that data control a car. \cite{tai_2016} use a deep learning approach to directly control an autonomous robot based on raw image data in a model-less approach.

Another use case of optical flow is the compression of videos. One codec that uses motion estimation is H.264/AVC \cite{ost_2004}. In the recent years, the requirements for video compression have steadily increased. Nowadays, many mobile devices are required to store and process video data efficiently. It is possible to extract the optical flow during the decompression of the video stream. Since the extraction is usually implemented in a GPU, the optical flow can be extracted with almost no CPU overhead from the raw video stream \cite{hol_2014}. The feasibility of this approach was demonstrated by implementing an optical flow based odometry sensor on a Raspberry Pi \cite{hei_2017}.

Until now, no attempts were made to utilize a deep learning approach in connection with optical flow in order to generate direct control instructions for an autonomous robot. The optical flow vector field contains different clues that can be used for collision avoidance, but it is difficult to describe an universal algorithm for collision avoidance that works under all circumstances.

Two major disadvantage of optical flow are that it only works in structured environments \cite{ber_1981}, which means that the proposed solution only works well in unstructured (outdoor) environments, and that the robot has to move at all times. Both is usually the case for \glspl{sar} applications and for flying robots.

An universal collision avoidance can possibly be implemented with a small set of training data compared to what would be required to directly learn a collision avoidance on full RGB or grayscale images. It is likely that using an optical flow vector field as input data leads to a neural network that can generalize better (as it works independently of camera settings and lightning conditions) and scales better to different use cases. The high information density of optical flow allow high frame rates. The vector data is freely available as a byproduct of camera video compression algorithms, allowing to deploy this novel procedure on low power embedded devices.

The goal of this work is to evaluate if such simple collision avoidance can be realized based on a machine learning approach. The result should be a universal solution that can be applied to different platforms and is demonstrated on a small racing car. The car should be fully controllable externally by a higher level instance such as a human driver or high level algorithms, but at the same time should be able to avoid collisions while driving. Therefore, the implemented solution can be seen as a proxy or a low level firewall (refer Fig. \ref{fig:system_classification}).

\tikzstyle{block}=[draw, minimum size=2em, align=center]
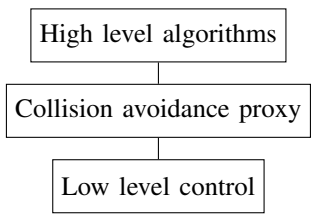
\begin{figure}[ht]
  \centering
  \begin{tikzpicture}[node distance=1cm,auto,>=latex']
      \node[block] (high_level) {High level algorithms};
      \node[block] (proxy) [below of=high_level] {Collision avoidance proxy};
      \node[block] (low_level) [below of=proxy] {Low level control};
      \path[-] (high_level) edge node {} (proxy);
      \path[-] (proxy) edge node {} (low_level);
  \end{tikzpicture}
  \caption{Classification of the implemented solution within the system}
  \label{fig:system_classification}
\end{figure}

\section{Method}

\subsection{Optical Flow in H.264/AVC}
H.264/AVC is a video coding standard that was introduced in 2003 to provide enhanced video compression for use cases like video streaming, requiring a high efficiency and data compression on error prone networks like UMTS or GSM \cite{ost_2004}.

The codec divides the image into macro blocks. For each macro block a motion compensation is predicted by estimating the displacement vectors to a reference frame, which is usually the last frame, but may also be a fixed frame. Each motion compensation macro block usually has a size of 16 px $\times$ 16 px. \cite{ost_2004}

The Raspberry Pi is a small single-board computer utilizing a 1 GHz single-core CPU, 512 MB RAM and the possibility to connect a camera \cite{ras_2019}. This camera compresses its video stream with H.264/AVC while the Raspberry Pi is decodes the video stream in its GPU in real time. The Raspberry Pi Foundation makes the motion vector stream available through the API \cite{hol_2014}.

\subsection{Testing Platform}
The goal of this work is to evaluate if optical flow can generally be used for collision avoidance tasks with the application in \glspl{mav} in mind. \glspl{mav} are expensive to develop and maintain and only allow a short consecutive operation due to energy constraints. Therefore, a small remote racing car was used as a testing platform. Even though the front-wheel steering and the rear-wheel drive (Ackerman steering configuration) results in a limited maneuverability \cite{sie_2004}, it behaves mostly like a fixed-wing \glspl{mav} or a bird flying in the air or rather very close to the ground, which makes it an attractive testing and evaluation platform.

In order to fully control the car, it is required to control the steering of the front wheels and the drive of the rear wheels. The car was shipped as a remote control car with an integrated battery, a servo motor for steering control and a brushed motor for driving, along with a remote receiver and transmitter. There were different options of transferring control information to the car:
\begin{enumerate}
  \item Keep the car as is and use the remote controller to feed steering and drive information wirelessly to the car
  \item Reverse engineer the remote receiver to directly feed control information to it via an embedded controller
  \item Replace the remote receiver with an embedded controller and add motor controllers for the servo motor and the drive motor
  \item Remove the servo motor and/or the drive motor and replace the whole electronics with a custom solution
\end{enumerate}

For a controller, different available solutions exist as well:
\begin{enumerate}
  \item Use an off-the-shelf controller like \cite{apm_2019} to interface the car with a high level controller
  \item Use a simple embedded custom micro controller to interface the car with a high level controller
  \item Directly control the car with a Raspberry Pi
\end{enumerate}

The modification of the car should take as little time as possible since this part is not the focus of this work. A Raspberry Pi must be integrated into the system for providing the optical flow and inferring the neural network. The Raspberry Pi is running Linux, which is not a real time operating system, why the car cannot be controlled directly by it. Apart from that, a Raspberry Pi has not a sufficient amount of \glspl{gpio}. The preferable solution would be to use a universal controller like ArduPilot, which is capable of controlling not only cars, but also other kind of vehicles like \glspl{mav} and can easily be controlled via a MAVLINK interface, therefore allowing to simply port the developed mechanisms to other platforms \cite{apm_2019}. Unfortunately, such controllers are relatively expensive. It was attempted to integrate a cheap clone of such a controller into the car, but the results did not look promising and the software was restricted to a closed source configuration tool, why a custom controller based on an Arduino \cite{dau_2012} was chosen.

Even though it would have been desirable to change as few things in the hardware of the car as possible, it was not possible to directly interface the integrated servo motor of the car. The servo motor consists of a single brushed motor and a sensor for measuring the current position of the steering wheels. Unfortunately, the servo controller was integrated into the receiver, which was removed from the car. To still be able to use the integrated servo motor, it would have been required to implement a controller into the replaced embedded controller of the car, which would mean an additional overhead and work. Therefore, the integrated servo motor was replaced with an off-the-shelf standard servo motor, which can be directly controlled with a \glspl{pwm} signal from the Arduino without any additional hardware.

The integrated drive motor did not have to be replaced, but required the use of a motor controller. Also, a speed regulation was desired, why a rotational encoder was added. In order to keep the system compatible to other controllers, it is desired that the drive motor can simply be controlled with a standard RC \glspl{pwm} signal like the steering servo. This again reduces the additionally required hardware to a minimum and allows to control the drive motor directly with the Arduino.

As mentioned in \ref{sec:introduction}, the optical flow does not allow to obtain absolute information about the environment. To have reproducible results, an additional measurement like a measured distance or a speed may have to be combined with the optical flow measurement. To accurately measure the speed, a magnet was glued to the motor's axis and a Hall-effect sensor placed close to it. This represents a closed loop with the Arduino for a simple speed regulation, which was implemented as a PI controller.

Furthermore, the car should be remote controlled. The removed remote control could not be integrated into the system. Therefore, a standard RC remote control was added, which the Arduino is capable of decoding via a \glspl{pwm}/\glspl{ppm} bridge.

The Arduino is now capable of measuring and controlling the speed of the car as well as set its steering. Also, it can read remote control parameters. For the Raspberry Pi to control the car, a communication to and from the Arduino was to be established. Both have a \glspl{uart}, which is used for this purpose. A simple communication protocol was implemented in Python and in C++.

Additionally, three ultrasonic sensors were added, which were not used in the final application.

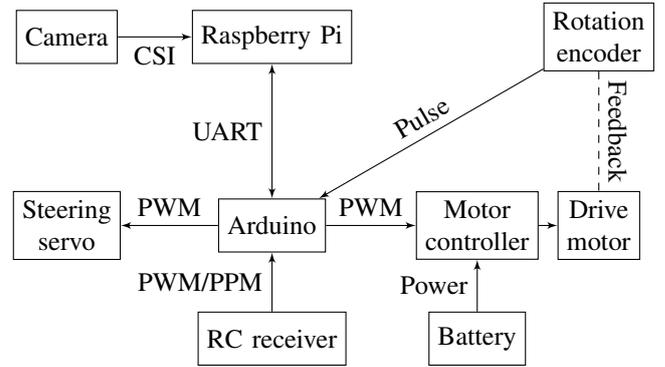
\begin{figure}[ht]
  \centering
  \begin{tikzpicture}[auto,>=latex']
    \node[block, node distance=2.7cm] (servo) {Steering\\ servo};
    \node[block, node distance=2.7cm] (arduino) [right of=servo] {Arduino};
    \node[block, node distance=2.5cm] (raspberry) [above of=arduino] {Raspberry Pi};
    \node[block, node distance=2.7cm] (camera) [left of=raspberry] {Camera};
    \node[block, node distance=1.5cm] (receiver) [below of=arduino] {RC receiver};
    \node[block, node distance=2.7cm] (mot_control) [right of=arduino] {Motor\\ controller};
    \node[block, node distance=1.5cm] (battery) [below of=mot_control] {Battery};
    \node[block, node distance=1.6cm] (motor) [right of=mot_control] {Drive\\ motor};
    \node[block, node distance=2.5cm] (encoder) [above of=motor] {Rotation\\ encoder};
    \path[->, above] (arduino) edge node {PWM} (servo);
    \path[<->] (arduino) edge node {UART} (raspberry);
    \path[<-] (raspberry) edge node {CSI} (camera);
    \path[->] (receiver) edge node {PWM/PPM} (arduino);
    \path[->] (arduino) edge node {PWM} (mot_control);
    \path[->] (mot_control) edge node {} (motor);
    \path[->] (battery) edge node {Power} (mot_control);
    \path[-, dashed, sloped, anchor=center, above] (encoder) edge node {Feedback} (motor);
    \path[->, sloped, anchor=center, above] (encoder) edge node {Pulse} (arduino);
  \end{tikzpicture}
  \caption{Block diagram of the car's electronics}
  \label{fig:car_blockdiagram}
\end{figure}

\begin{figure*}[ht]
  \centering
  \begin{minipage}[b]{0.24\textwidth}
    \centering
    \includegraphics[width=\linewidth]{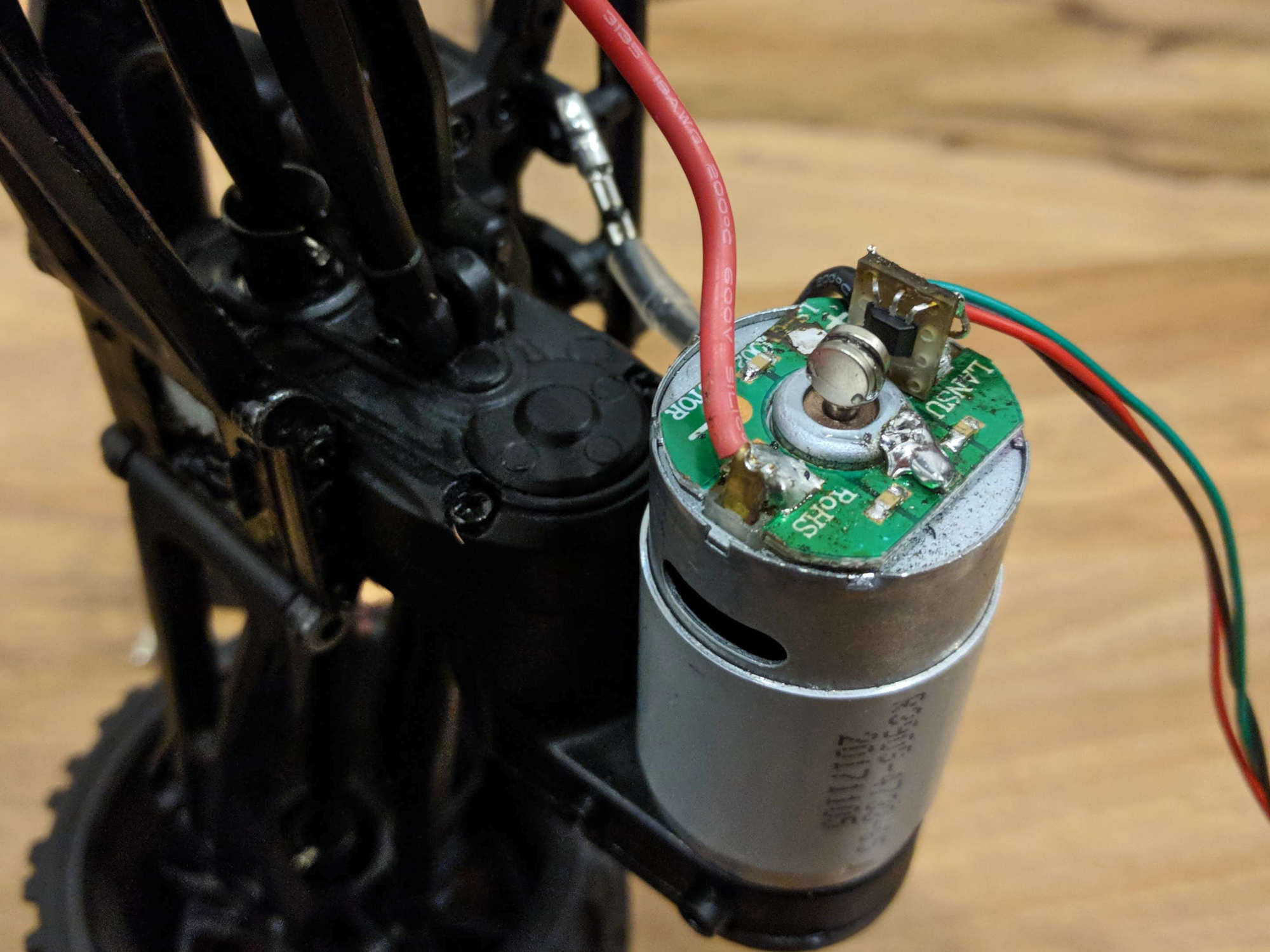}
    \subcaption{Drive motor with magnets attached to the shaft and incremental encoder (Hall sensor)}
  \end{minipage}
  \hfil
  \begin{minipage}[b]{0.24\textwidth}
    \centering
    \includegraphics[width=\linewidth]{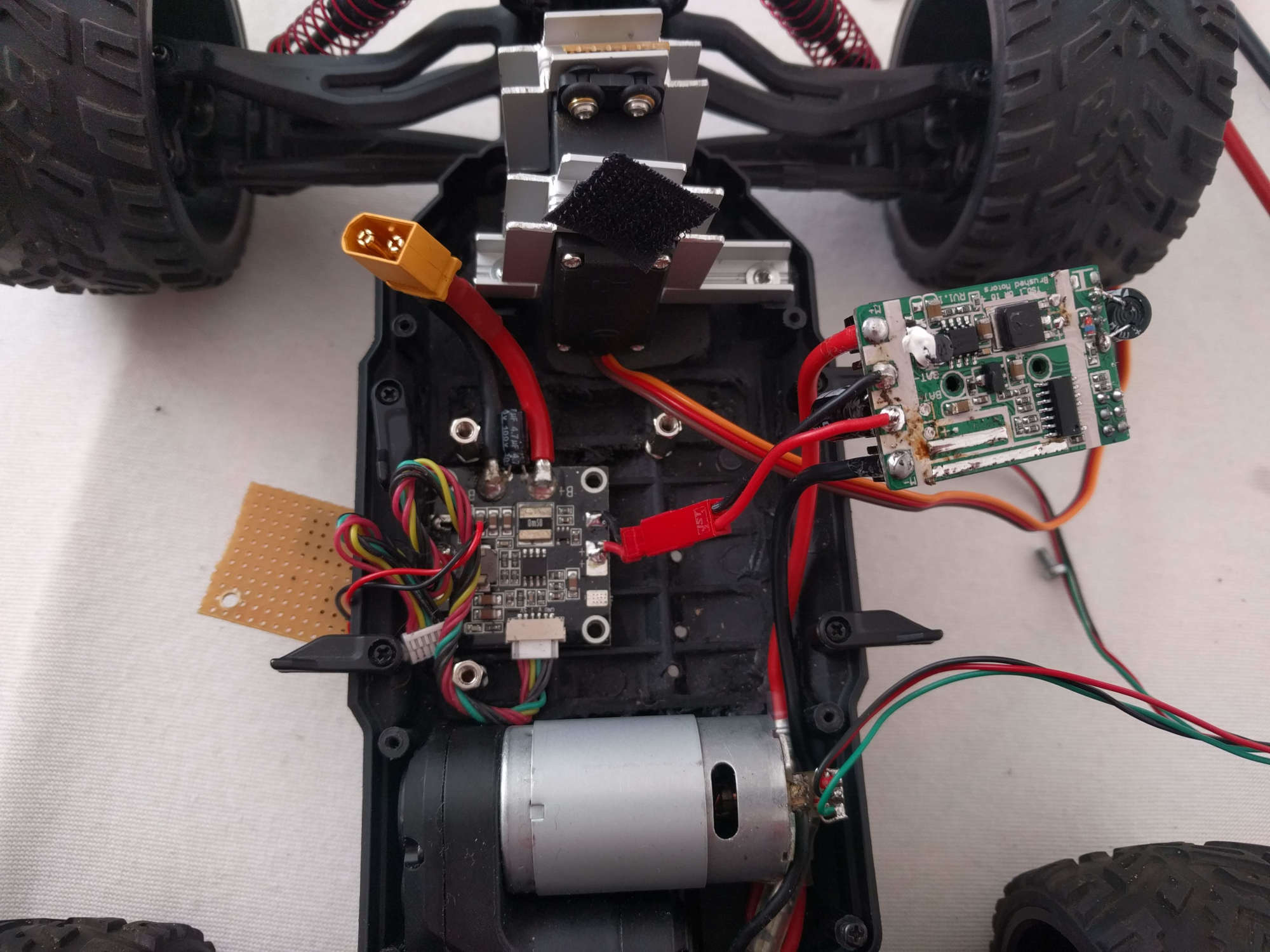}
    \subcaption{Power train and steering unit: Steering servo, motor controller, power distribution, drive motor}
  \end{minipage}
  \hfil
  \begin{minipage}[b]{0.24\textwidth}
    \centering
    \includegraphics[width=\linewidth]{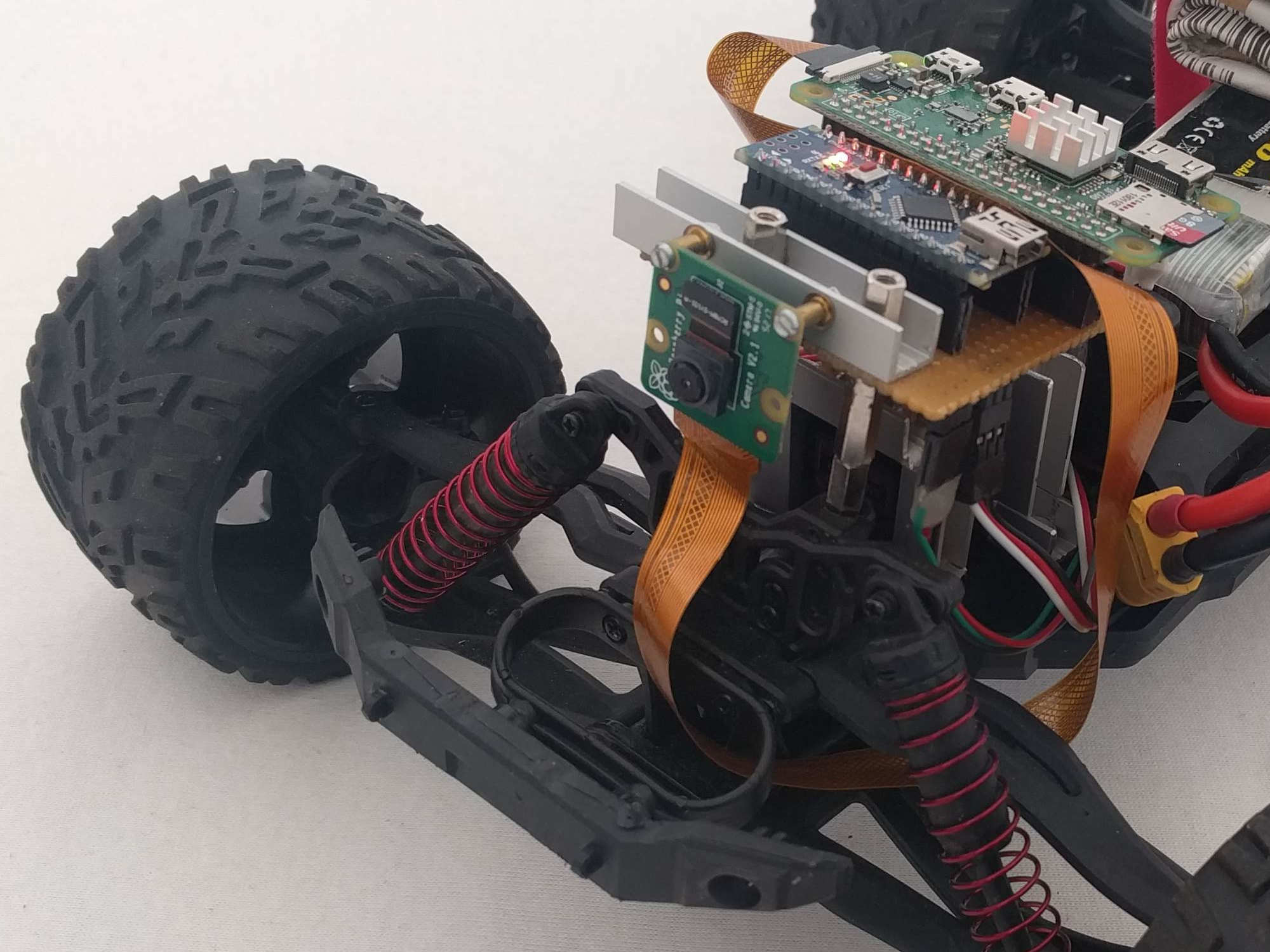}
    \subcaption{Frontal view with the camera, Arduino and Raspberry Pi CPU and the battery in the background}
  \end{minipage}
  \hfil
  \begin{minipage}[b]{0.24\textwidth}
    \centering
    \includegraphics[width=\linewidth]{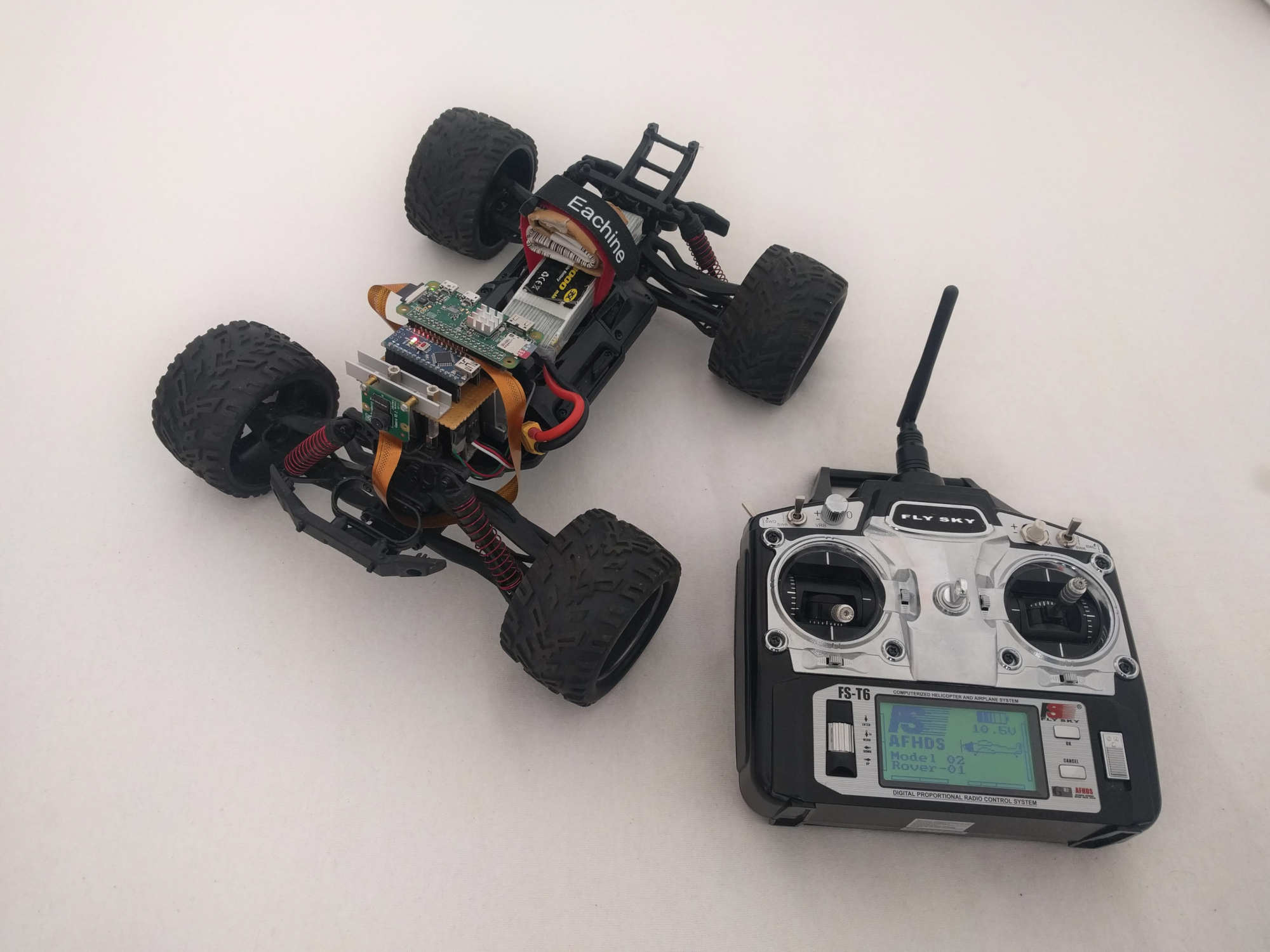}
    \subcaption{Complete setup: Modified car with remote controller\\}
  \end{minipage}
  \caption{Components and development process of testing platform}
  \label{fig:car_overview}
\end{figure*}

The whole system is visualized in the block diagram in Fig. \ref{fig:car_blockdiagram} and close-ups of the different components and the whole system are depicted in Fig. \ref{fig:car_overview}.

\subsection{Machine Learning}

While humans are good at solving seemingly simple problems like speech recognition and detecting faces and bad in solving tasks that require large computations, it is the other way around for computers, which are good at solving tasks that can be described by a set of formal rules. The capability of automatically extracting patterns from raw data and thus acquiring knowledge to solve a certain problem, which are often considered intuitive by humans, is known as machine learning. \cite{goo_2016}

The performance of machine learning approaches depends on the representation of the data to analyze. Machine learning can either be applied to a set of defined features or even detect features on its own and then use the detected features to perform a further classification. Often, the approaches based on automatically extracted features perform better than approaches based on manually selected features, since it is often difficult to describe a general set of features for a complex problem. Features often depend on other features and conditions. Deep learning attempts to solve this problem by automatically breaking a problem down into features that are described by other features. \cite{goo_2016}

Machine learning is a mean to solve a certain problem, or a task. Two of the most common tasks are classification and regression. In a classification task the machine is supposed to determine which of $n$ classes a given input is to be assigned to, i.e. finding a solution to the function $f : \mathbb{R}^n \rightarrow \{1, ..., n\}$. It is also possible for $f$ to output the probability distribution over a set of $n$ classes. A common use case for a classification task is object detection. In a regression task the machine is supposed to predict a value, i.e. finding a solution to the function $f : \mathbb{R}^n \rightarrow \mathbb{R}$, which is quite similar to the classification. An example is to find the position of an object in an image. \cite{goo_2016}

Machine learning can be categorized in supervised and unsupervised learning. Unsupervised learning occurs on a data set without labels and aims to extract patterns like a probability distribution or finding clusters. Supervised learning on the other hand occurs on a labeled data set with the goal to understand the connection between a data point and a given label. \cite{goo_2016}

Deep Feedforward Networks (DNNs) are what is commonly known as neural networks. DNNs consist of multiple layers, or vector to vector functions $f$, which are composed so that the information flows from one direction to the other. At this point, the network is still linear. By introducing activation functions, nonlinearities are integrated into the network. The multiple functions, or layers, have neurans with weights which must be determined by the machine. For this, backpropagation is used to find the overall derivative function of a model. Based on that and a loss function, a gradient descent can be executed to find a minimum in the via the backpropagation determined function and therefore finding the solution to the problem. During the learning phase, the weights are adapted accordingly during multiple epochs, during which the whole data set is considered for the learning each cycle. \cite{goo_2016}

There are multiple kinds of layers in a network. The simplest layer is often referred to as fully connected layer and is applied by a simple vector-matrix multiplication (where the vector are the values of the output of the last layer and the matrix contains the weights). Another important type of layers are convolutional layer, which are basically the same as fully connected layers, but applied to a defined region of data. In a two-dimensional image, a convolution may have a filter size of three by three pixels and move in a grid over the two image. Pooling layers downsample a layer. \cite{goo_2016}

The goal for all machine learning approaches is to work well on unseen data, i.e. to generalize. The learning process involves a training and a testing data set. While the network uses the training data to minimize a loss function, i.e. an error, the testing data can be used to determine how well the network generalizes. The goal is not only to minimize the error on the training set, but also to keep the difference between training and testing error small. During the learning process, the gap between training and testing error usually initially decreases and converges. During this time, the network is underfitted. At some point, the two curves might diverge again, i.e. the gap between training and testing error diverges. After this point, the network is overfitting. \cite{goo_2016}

For this work, deep convolutional networks are used. One particularly well-known classification network is VGG \cite{sim_2014}. VGG exists in multiple configuration, the smallest consisting of eleven layers, where the first eight are convolutions and pooling layers, while the last three are fully connected layers. Intuitively, the first layers extract features from the input image from a low level to a very high level and the fully connected layers then determine, based on the features found, what object is seen.

\subsection{Data collection}
The basis for all machine learning solutions is a sufficiently large and statistically appropriate data set \cite{goo_2016}. In this work, the goal for the robot is to learn to avoid collisions based on gathered training data. There are different approaches for gathering this data, which were considered iteratively and are summarized in this paragraph.

The simplest solution is to record optical flow data with the car while driving and later labeling the images manually. The problem with this approach is that a large amount of time has to be invested into manually labeling the images recorded initially. A solution in which the labeling can be performed automatically is to be preferred. This solution was still pursued due to different weaknesses of the other approaches.

The second approach is to add distance sensors to the robot and record distance measurements synchronized to the optical flow. Something similar was already done, but with a simple monocular camera instead of optical flow \cite{mic_2005}. It was attempted to mount three ultrasonic distance sensors on top of the car, but the produced measurements were unreliable, likely due to the vibrations of the car and due to the limited update rate. Another option was to add a \glspl{lidar} to the car. While mapping distance values to the optical flow would be an interesting task, this would go beyond the scope of this work and would not be an end-to-end solution, why this idea was discarded.

The last approach is the end-to-end approach where the movement of the car should be controlled completely by a neural network. Something similar was done in \cite{yan_2005}, but with a normal camera. Initially, only the desired steering of the car was recorded. The idea was that the car learns based on this data on its own what desired maneuvers are and what approaching obstacles are. According to \cite{yan_2005} this generally works, but a vast amount of data has to be recorded and the operator of the car has to be consequent on their obstacle avoidance decisions. There, the steering decisions of the operator were recorded together with the video data. The operator wore video googles and was able to see the recordings of the car in real time. The vehicle had to drive straight ahead if no obstacle was seen and consistently drive left or right along an obstacle if one occurred. \cite{yan_2005}

\begin{figure*}[ht]
  \centering
  \begin{minipage}[b]{0.22\textwidth}
    \centering
    \includegraphics[width=\linewidth]{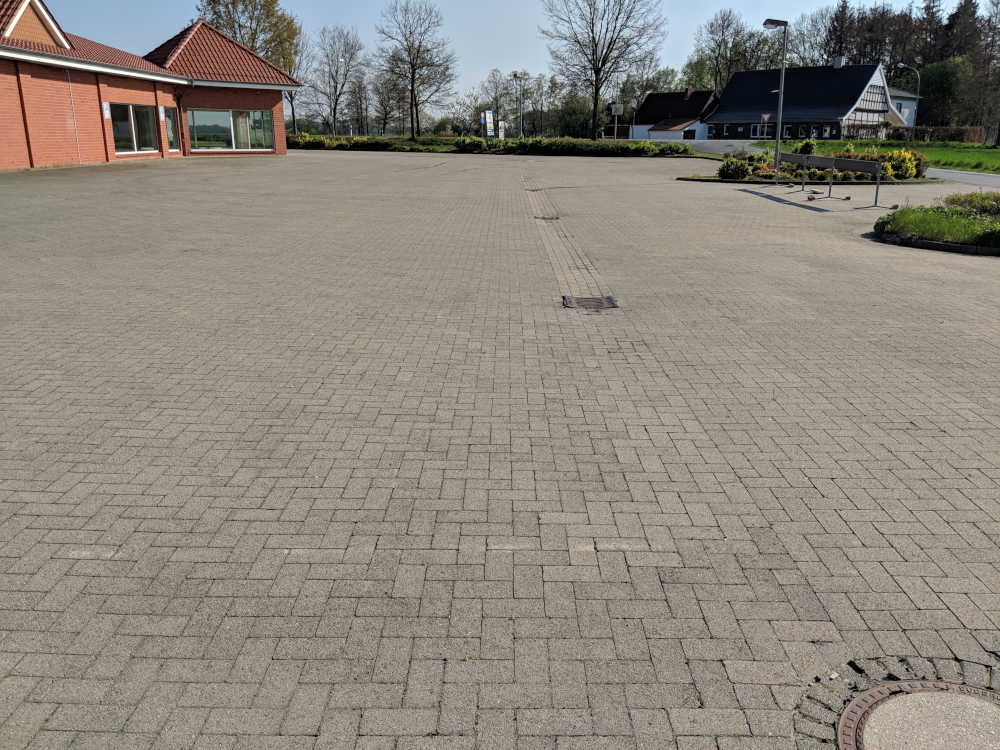}
    \subcaption{\textbf{Parking spot:} A parking spot limited by a building and border stones, sometimes with small vegetation. The car was driven once in both directions around the perimeter. This represents collisions with obstacles on the front left or front right side of the car.}
  \end{minipage}
  \hfil
  \begin{minipage}[b]{0.22\textwidth}
    \centering
    \includegraphics[width=\linewidth]{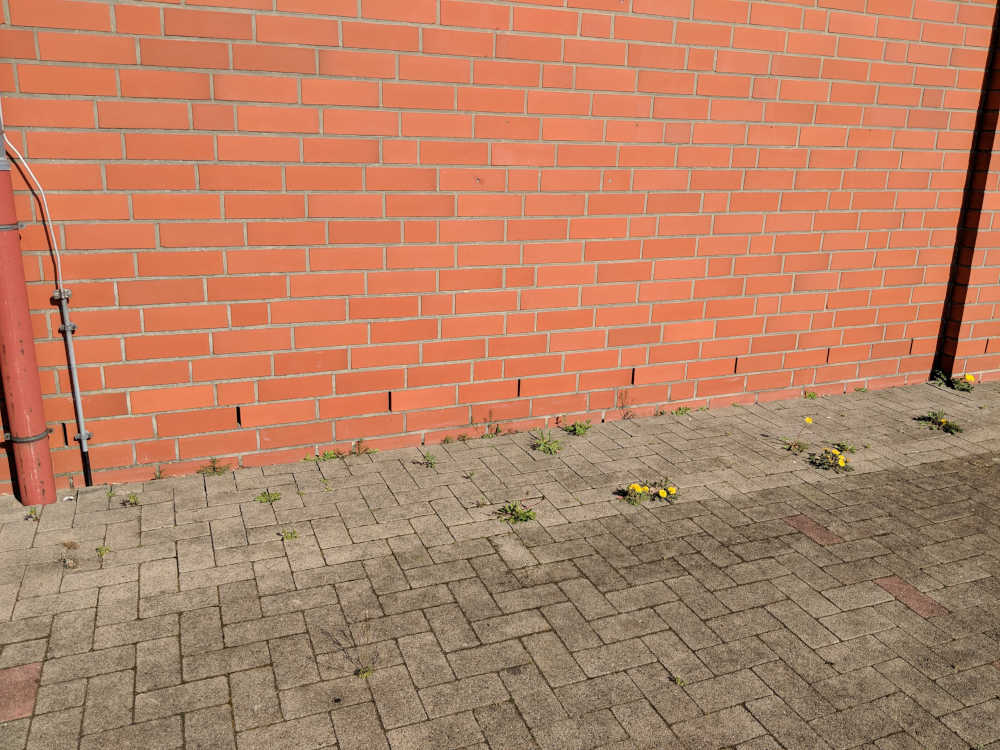}
    \subcaption{\textbf{Frontal collision:} The car was steered continuously frontal towards a wall of the building on the parking spots at different angles and then driven back to the starting position. This data set represents large structured obstacles that are approached almost frontally.}
  \end{minipage}
  \hfil
  \begin{minipage}[b]{0.22\textwidth}
    \centering
    \includegraphics[width=\linewidth]{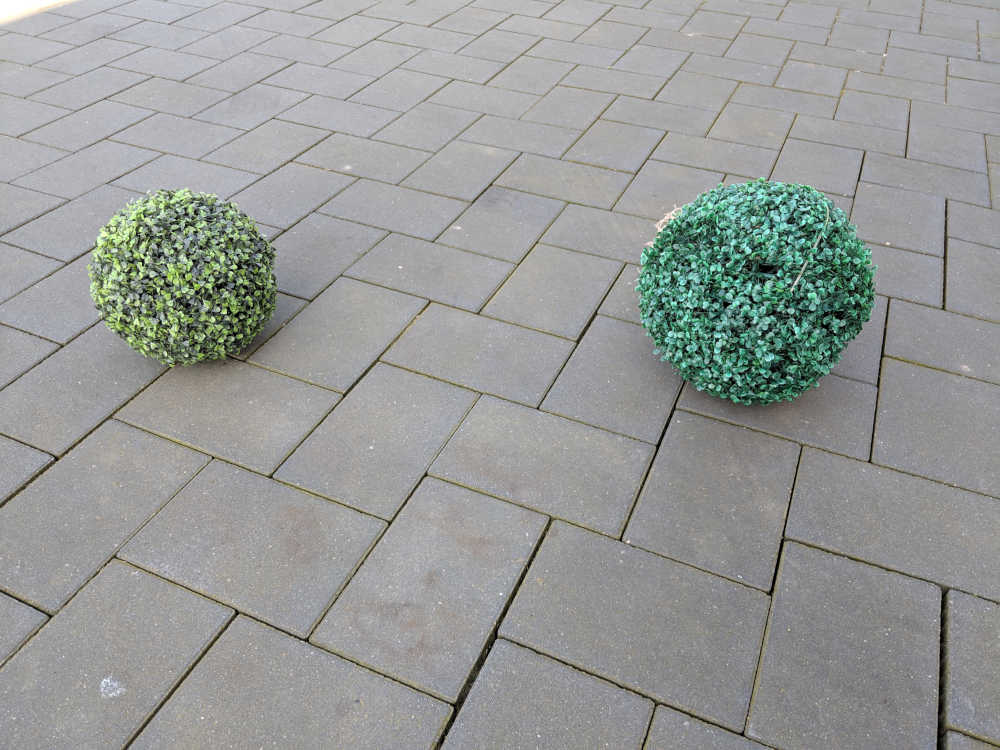}
    \subcaption{\textbf{Spherical obstacle:} Two spherical obstacles were placed on an even underground. The underground is shadowy due to a building on one site, while the background is sunny, offering a large dynamical contrast. Two cars are parked close by.}
  \end{minipage}
  \hfil
  \begin{minipage}[b]{0.22\textwidth}
    \centering
    \includegraphics[width=\linewidth]{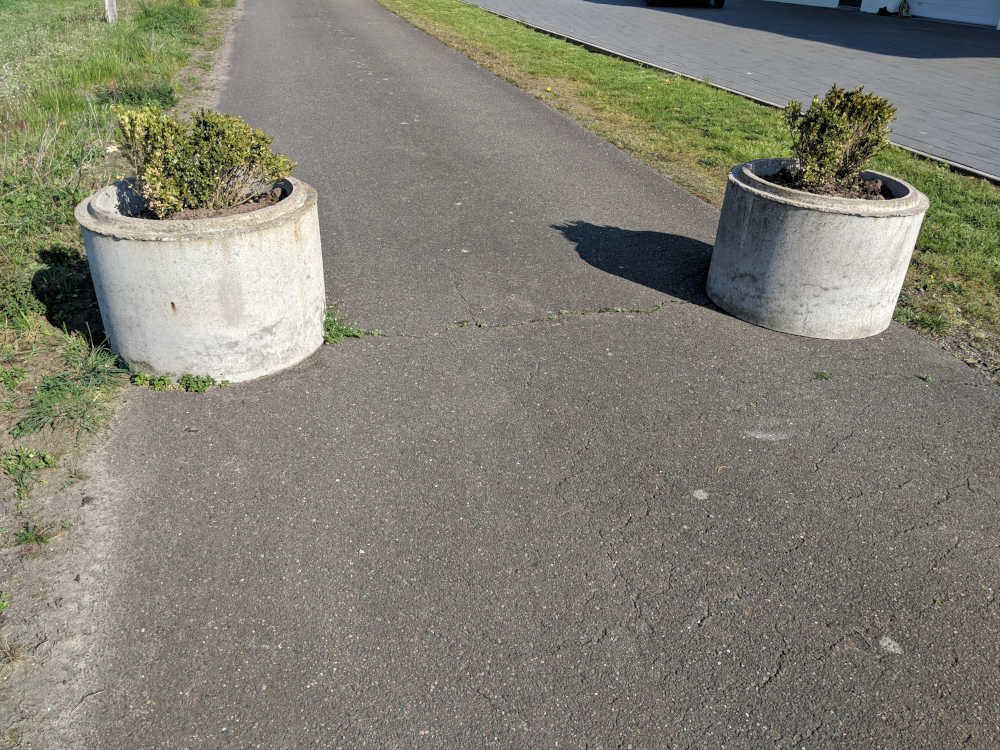}
    \subcaption{\textbf{Cylindrical obstacle:} Two poller block a street with some room in between. The car was steered towards and between the pollers so that the car should not only be capable of detecting and avoiding obstacle, but also to learn if there is sufficiently space.}
  \end{minipage}
  \caption{Data set locations}
  \label{fig:dataset_locations}
\end{figure*}

It would be desirably to perform a similar method of data collection for this work, but there are two primary restrictions that did not allow a similar procedure. Firstly, the testing platform that was used with this work is not capable for driving off-road due to its size. It might be possible to make the car capable for off-road driving, but this works goal is to prove the general functionality of machine learning based optical flow collision avoidance, so a simple on-road obstacle avoidance demonstration would suffice. Secondly, a live video stream was not possible to realize due to time and budget restrictions. Instead, the car operator followed the car and avoided collisions from his point of view. The data recordings primarily took place on a large car parking spot with a building on one site and vegetation and border stones on the other side (refer Fig. \ref{fig:dataset_locations}).

Recorded were the optical flow, the current driving speed of the car and the remote control signals. Since the car should not only randomly drive around and avoid obstacle, but follow a desired path and avoid obstacles it encounters, a desired path was recorded with one control signal and a corrected path was recorded with a second control signal. The operator determined a desired path. As soon as the operator perceived a possible collision, a correction signal was sent to the car with a second control signal that would override the first signal and correct the driven path of the car.

\begin{figure*}
  \centering
  \includegraphics[width=0.9\textwidth]{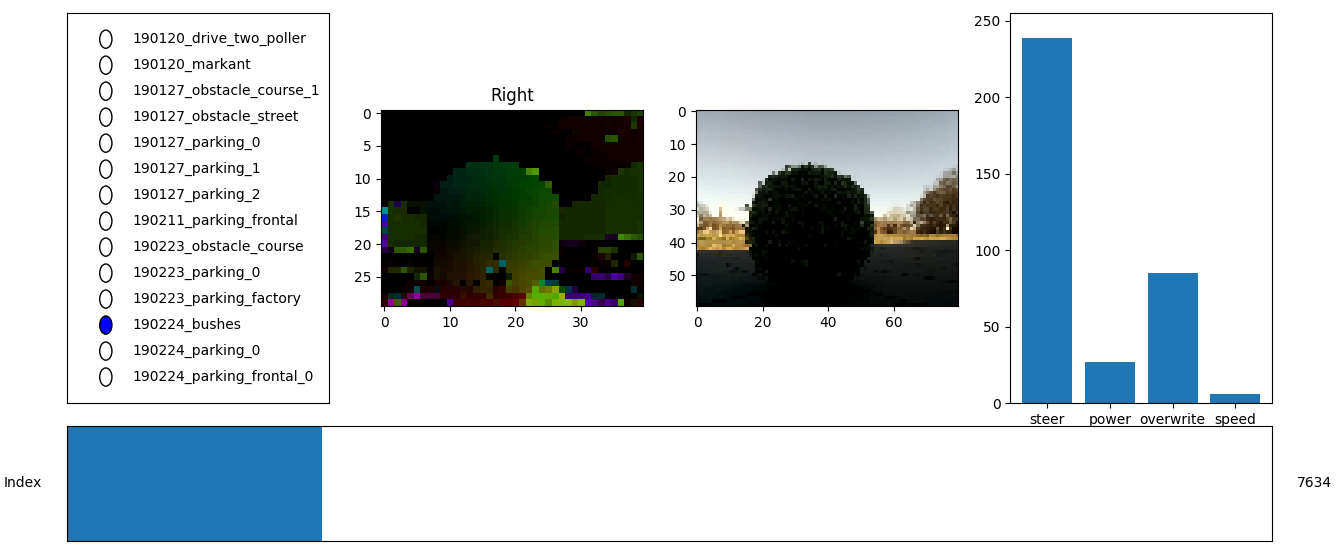}
  \caption{Visualization and labeling tool for data sets. On the left side the available data sets are listed. Next to the list the currently selected frame as the optical flow and the raw RGB view are visible. The current frame can be selected either with the slider in the bottom or with shortcuts. On the right side the steering signal and the speed of the car is visualized. The label that is assigned to this image is drawn above the optical flow image. Labels can be changed with key shortcuts as well. Since the data in the data sets is stored in sequence of the recording, multiple frames can be labeled at once since usually the same action is performed in consecutive images. This way, almost 80k frames were labeled in a few hours.}
  \label{fig:labeling_tool}
\end{figure*}

During later recording sessions, the raw video stream of the camera was recorded additionally to the optical flow. Since the operator did not see what the car has seen, the correction signals sent by the operator are not consistent to the video stream, why a manual labeling of the data was performed additionally. In order to speed up the manual labeling, only three classes were assigned to the data, which are 1) no correction (drive as desired), 2) fully steer left 3) fully steer right. This way, the car would not provide a continuous correction signal, but fully steer in the opposite direction of the obstacle, which is sufficient for a proof of concept. The developed labeling tool can is depicted in Fig. \ref{fig:labeling_tool}.

\begin{figure*}
  \centering
  \begin{minipage}[b]{\textwidth}
    \centering
    \begin{tabular}{cccc}
      \includegraphics[width=0.22\textwidth]{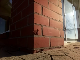} &   \includegraphics[width=0.22\textwidth]{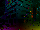} & \includegraphics[width=0.22\textwidth]{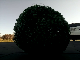} &   \includegraphics[width=0.22\textwidth]{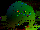} \\
      \includegraphics[width=0.22\textwidth]{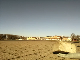} &   \includegraphics[width=0.22\textwidth]{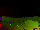} & \includegraphics[width=0.22\textwidth]{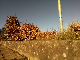} &   \includegraphics[width=0.22\textwidth]{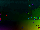}
    \end{tabular}
    \subcaption{Example frames for which obstacles can clearly be identified in the optical flow \newline}
  \end{minipage}
  \hfil
  \begin{minipage}[b]{\textwidth}
    \centering
    \begin{tabular}{cccc}
      \includegraphics[width=0.22\textwidth]{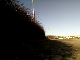} &   \includegraphics[width=0.22\textwidth]{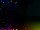} & \includegraphics[width=0.22\textwidth]{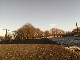} &   \includegraphics[width=0.22\textwidth]{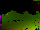} \\
      \includegraphics[width=0.22\textwidth]{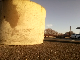} &   \includegraphics[width=0.22\textwidth]{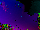} & \includegraphics[width=0.22\textwidth]{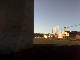} &   \includegraphics[width=0.22\textwidth]{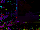}
    \end{tabular}
    \subcaption{Example frames for which obstacles can not be clearly identified due to bad lighting conditions or when the background appears to be an obstacle}
  \end{minipage}
  \hfil
  \caption{Positive and negative examples for the collected optical flow data. The optical flow is encoded in the HSV color space where the hue encodes the direction of the flow and the value the speed. The RGB images are stored in a low resolution to avoid wasting memory.}
  \label{fig:flow_examples}
\end{figure*}

Ultimately, three data sets were recorded with raw RGB images, optical flow, desired steering, corrected steering, actual speed and additionally manually labeled into the classes left, straight, right, resulting in n frames. The car was driven around by controlling the desired steering signal. When an obstacle was approached, the override signal was triggered by the human operator.

A few sample images of the data set can be seen in Fig. \ref{fig:flow_examples}. From the images it becomes clear that obstacle features are easily detectable by a human perspective, sometimes even better than in the plain RGB image. The images show a resemblance to depth maps.

\subsection{Neural Network Evolution}
The neural network which is to be implemented should ideally be capable of outputting a steering signal which follows the desired steering signal by the higher level algorithms and is corrected if an obstacle is approached. The approaches that were implemented in this work considered a regression as well as a classification approach (i.e. obstacle is seen left/right).

The neural networks are implemented in Python with the Keras Library and Tensorflow as backend.

One of many image classification networks is \cite{sim_2014}. This network architecture extracts features from an image with multiple layers of convolutions and feeds these features into a fully connected network, which then determines classes. This network is the base for both main approaches of this work. While the original VGG network has 138 million trainable parameters \cite{sim_2014}, the networks used in this work are required to have parameters in the order of less than a few thousand parameters due to computation power constraints on the Raspberry Pi.

The base architecture used in both approaches is described in Table \ref{tab:general_architecture}. The input tensor is the optical flow image. The first convolution has 32 filters of the size 3x3. A max-pooling layer with the kernel size 2x2 reduces the size by the factor of four, followed by two convolutions each with a 3x3 kernel and 8 filters each. A second max-pooling layer reduces the tensor again, before it is fed to two fully connected layers with 16 neurons each and one output neuron. This base architecture has 6601 parameters and is the base architecture for the two main approaches.

\begin{table}
  \begin{tabular}{l|l|l|l}
    Type  & Shape        & Parameters & Description                             \\
    \hline
    Input & (30, 40, 2)  & 0          & \begin{tabular}[c]{@{}l@{}}Optical flow input layer x, y,\\ (dx, dy)\end{tabular} \\
    Conv  & (28, 38, 32) & 608        & 2D Convolutions, kernel (3, 3)          \\
    Pool  & (14, 19, 32) & 0          & Max-Pooling, kernel (2, 2)              \\
    Conv  & (12, 17, 8)  & 2312       & 2D Convolutions, kernel (3, 3)          \\
    Conv  & (10, 15, 8)  & 584        & 2D Convolutions, kernel (3, 3)          \\
    Pool  & (5, 7, 8)    & 0          & Max-Pooling, kernel (2, 2)              \\
    Dense & 16           & 4496       &                                         \\
    Dense & 16           & 272        &                                         \\
    Dense & 1            & 17         &
  \end{tabular}
  \caption{Description of the overall layer setup (6601 Parameters)}
  \label{tab:general_architecture}
\end{table}

\subsubsection{Regression approach}
For the regression approach, the neural network is fed with the optical flow vector field as well as the desired steering. The optical flow image is processed by different layers of convolutions in order to extract features. At the end of the convolution layers, multiple fully connected layers convert extracted features to an output signal. The desired steering signal is connected to the first fully connected layer.

\tikzstyle{init} = [pin edge={to-,thin,black}]
\begin{figure}[ht]
  \centering
  \begin{tikzpicture}[node distance=2.7cm,auto,>=latex']
    \node[block] (conv) {Convolutions};
    \node (start) [left of=conv, coordinate] {};
    \node[block, pin={[init]above:Desired Steer}] (dense) [right of=conv] {Fully Connected};
    \node (end) [right of=dense, coordinate] {};
    \path[->] (start) edge node {Image} (conv);
    \path[->] (conv) edge node {} (dense);
    \path[->] (dense) edge node {Steer} (end);
  \end{tikzpicture}
  \caption{Overview of the regression architecture}
\end{figure}
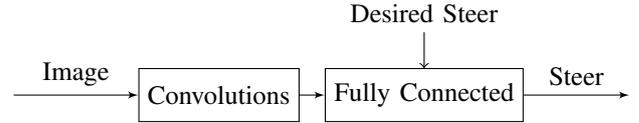

The training data for the regression approach was not labeled manually, since a manual labeling of a continuous output signal is not possible. Instead, the desired steering signal is compared with the corrected steering signal in the training data. If the corrected steering signal is not triggered, the desired steering signal is marked as the label for the corresponding optical flow frame. If the corrected steering signal is triggered (i.e. the operator of the manual determined a collision was about to occur), the corrected steering signal was used as the label for the frame.

The data set that is generated with this strategy contains 51887 samples which can be split into the two cases where the desired steering is passed through the network when no obstacle is seen (11742 samples) and the case where the desired steering is changed in order to avoid obstacles (40145). It can be seen that these two cases are unequally large, or not balanced. Unbalanced training data likely, but not necessarily, influences the performance of the neural network and results in the network having difficulties learning the outnumbered class \cite{bat_2004}. This can either be resolved by recording more training data covering the first case or by discarding training data of the second case. This procedure is called data balancing \cite{bat_2004}. After balancing, the data set consists of 11742*2 = 23484 samples. In retrospect, it was found that Keras has an option for weighting different classes so that valuable training data would not have to be discarded, but this only works for a classification approach.

Since this network is based on a regression approach, the output neuron has a linear activation function. The loss function for this network is a simple mean-squared-error, which means that the networks tries to minimize the squared difference between the labeled steering and the predicted steering. The data sets were split into a training (80\%) and a test set (20\%).

After 22 epochs, the network converges with a loss decreased to 0.0116 for the unbalanced data and after 54 epochs with a loss of 0.0096 for the balanced data.

\newlength\figurewidth
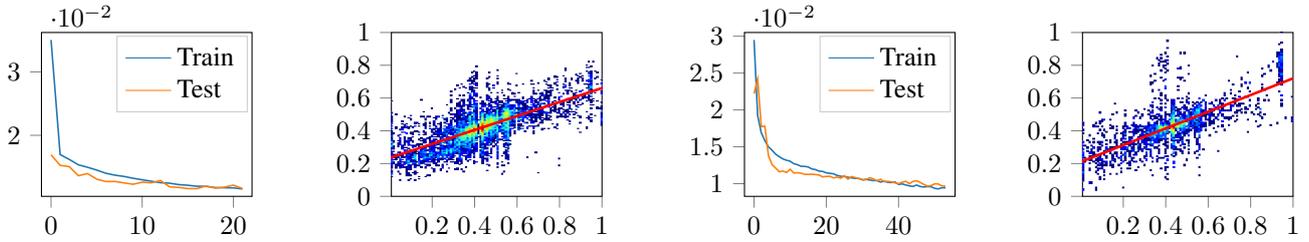
\begin{figure*}
  \centering
  \begin{minipage}[b]{0.24\textwidth}
    \centering
    \setlength\figurewidth{\linewidth}
    \input{img/plots/regression_training_loss_unbalanced.tex}
    \subcaption{Loss/Epochs for unbalanced data. Final test loss after 22 epochs: 0.0116}
    \label{fig:regr_train_loss_unbalanced}
  \end{minipage}
  \hfil
  \begin{minipage}[b]{0.24\textwidth}
    \centering
    \setlength\figurewidth{\linewidth}
    \input{img/plots/regression_evaluation_unbalanced.tex}
    \subcaption{Histogram for unbalanced data. Best fit slope/interception: 0.42, 0.23}
    \label{fig:regr_eval_unbalanced}
  \end{minipage}
  \hfil
  \begin{minipage}[b]{0.24\textwidth}
    \centering
    \setlength\figurewidth{\linewidth}
    \input{img/plots/regression_training_loss_balanced.tex}
    \subcaption{Loss/Epochs for balanced data. Final test loss after 54 epochs: 0.0096}
  \end{minipage}
  \hfil
  \begin{minipage}[b]{0.24\textwidth}
    \centering
    \setlength\figurewidth{\linewidth}
    \input{img/plots/regression_evaluation_balanced.tex}
    \subcaption{Histogram for balanced data. Best fit slope/interception: 0.51, 0.21}
  \end{minipage}
  \caption{Comparison of learning process and predictions in the regression approach with balanced vs. unbalanced data.}
  \label{fig:eval_regression}
\end{figure*}

\subsubsection{Classification approach}

For the classification approach, the neural network is solely fed with the optical flow vector field as well. The optical flow image is processed by different layers of convolutions in order to extract features. At the end of the convolution layers, multiple fully connected layers convert extracted features to a classification of the current scenario in either fully override steering left (meaning an obstacle is approaching frontal or right), fully override steering right (meaning an obstacle is approaching frontal or left) or passing through the desired steering (meaning no obstacle is approaching). Therefore, the last layer in the base architecture of Table \ref{tab:general_architecture} receives two additional neurons.

Since this approach is a multi class classification problem, the labels have to be converted to a one-hot scheme. Additionally to the mean-squared-error loss the accuracy is added and monitored as a metric. The output layer of the network has a softmax activation function, which means that three numerical values between 0 and 1 are calculated which sum to 1. The accuracy describes how often a maximum in the prediction vectors is equal to the label.

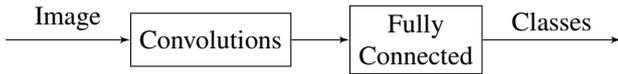
\begin{figure}[ht]
  \centering
  \begin{tikzpicture}[node distance=2.7cm,auto,>=latex']
    \node[block] (conv) {Convolutions};
    \node (start) [left of=conv, coordinate] {};
    \node[block] (dense) [right of=conv] {Fully\\ Connected};
    \node (end) [right of=dense, coordinate] {};
    \path[->] (start) edge node {Image} (conv);
    \path[->] (conv) edge node {} (dense);
    \path[->] (dense) edge node {Classes} (end);
  \end{tikzpicture}
  \caption{Overview of the classification architecture}
\end{figure}

Just like with the regression approach, the labeled data can be generated automatically depending on the desired steering and the corrected steering. For the classification approach, it is significantly easier to manually label the frames. Therefore, the classification approach was attempted and compared with automatically labeled data as well as manually labeled data with both balanced and unbalanced data.

\begin{table}[ht]
  \begin{tabular}{l|l|l|l|l|l}
                          & Epochs & Overall & Left  & None  & Right \\ \hline
    Automatic, Unbalanced & 52     & 73.25   & 69.69 & 76.97 & 72.67 \\
    Automatic, Balanced   & 46     & 72.15   & 68.12 & 79.24 & 68.93 \\
    Manual, Unbalanced    & 64     & 90.15   & 59.48 & 96.66 & 60.77 \\
    Manual, Balanced      & 56     & 79.64   & 83.59 & 70.32 & 85.37
  \end{tabular}
  \caption{Comparison of amount of epochs and accuracy (percentage) of the classification approaches}
  \label{tab:classification_accuracy}
\end{table}

Table \ref{tab:classification_accuracy} clearly shows that the combination of manual labels and balanced data shows the best overall results. This combination was used to analyze and optimize the network structure by variying single parameters while still staying below around 10000 trainable network parameters (refer Table \ref{tab:effect_manipulate_layers}).

\begin{table*}
    \centering
    \begin{tabular}{p{8cm}llllll}
                                                                                                                                & Parameters & Epochs & Overall & Left  & None  & Right \\
    Base architecture                                                                                                           & 8323       & 56     & 79.64   & 83.59 & 70.32 & 85.37 \\
    Three convolutions with 16 filters each in first layer                                                                      & 10099      & 59     & 80.8    & 83.9  & 70.32 & 88.55 \\
    Five convolutions with 8 filters each in second layer                                                                       & 6619       & 60     & 77.23   & 80.13 & 67.18 & 84.82 \\
    Increase the amount of neurons to 32 in the first dense layer                                                               & 13075      & 71     & 82.01   & 82.93 & 75.55 & 87.86 \\
    Increase the amount of neurons to 32 in the second dense layer                                                              & 8643       & 61     & 80.0    & 81.33 & 76.6  & 82.2  \\
    Adding a convolution with 16 filters in the first layer                                                                     & 12627      & 60     & 78.93   & 77.99 & 78.56 & 80.27  \\
    Decreasing the amount of filters in the first convolution to 8                                                              & 6139       & 91     & 80.13   & 78.66 & 80.13 & 81.65 \\
    Decrease the amount of filters in the first convolution to 8 and remove the second convolutions                             & 8555       & 67     & 78.39   & 83.86 & 69.41 & 82.2
    \end{tabular}
    \caption{Effect to the performance of the network by manipulating single layers: Amount of parameters, epochs until convergence and accuracy}
    \label{tab:effect_manipulate_layers}
\end{table*}

For the given task, the information that can be derived from the optical flow is still quite dense. Therefore, it was additionally evaluated how different input masks applied to the raw dense optical flow field with a size of 30 by 40 vectors affect the performance of the network. This evaluation can be seen in Table \ref{tab:different_masks}.

\begin{table}
  \begin{tabular}{llllllll}
    \rotatebox{60}{Shape}       & \rotatebox{60}{Mask}                    & \rotatebox{60}{Params} & \rotatebox{60}{Epochs} & \rotatebox{60}{Overall} & \rotatebox{60}{Left}  & \rotatebox{60}{None}  & \rotatebox{60}{Right} \\
    30, 40 & \includegraphics[width=1cm, frame]{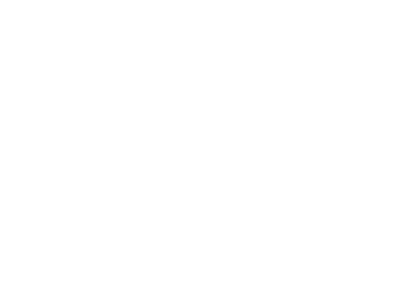}  & 8323       & 56     & 79.64   & 83.59 & 70.32 & 85.37 \\
    30, 20 & \includegraphics[width=1cm, frame]{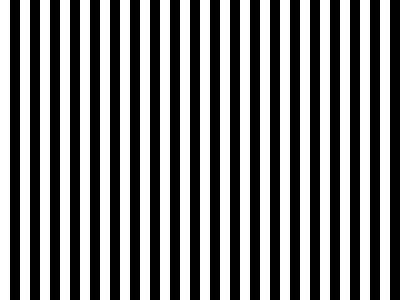}  & 5123       & 70     & 77.32   & 69.46 & 79.34 & 83.31 \\
    15, 40 & \includegraphics[width=1cm, frame]{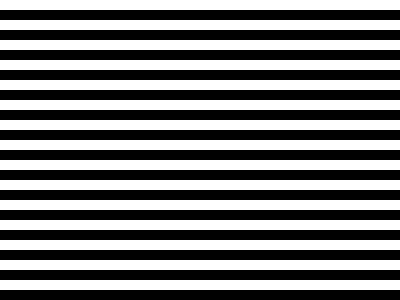}  & 4739       & 77     & 79.33   & 68.13 & 83.26 & 86.75 \\
    15, 20 & \includegraphics[width=1cm, frame]{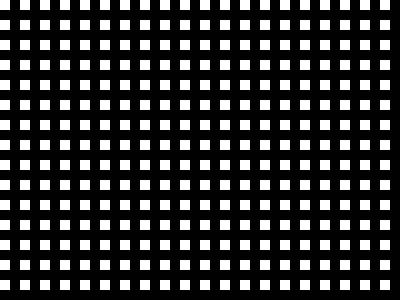}  & 4099       & 70     & 75.54   & 70.26 & 71.5  & 85.24 \\
    15, 40 & \includegraphics[width=1cm, frame]{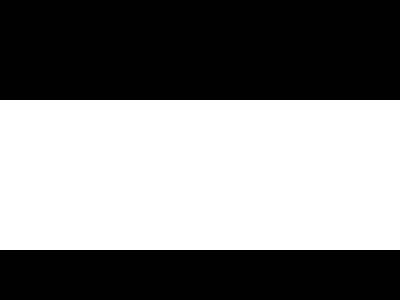} & 4739       & 83     & 79.15   & 74.8  & 79.6  & 83.17 \\
    15, 20 & \includegraphics[width=1cm, frame]{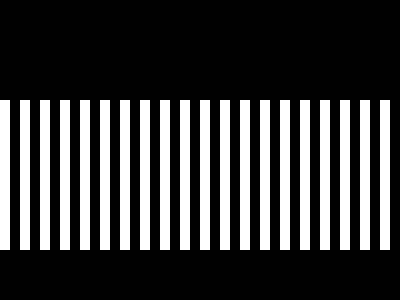}  & 6403       & 57     & 79.37   & 75.59 & 81.43  & 81.1 \\
    5, 40  & \includegraphics[width=1cm, frame]{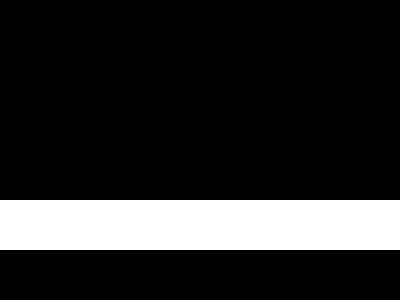} & 5123       & 81     & 74.24   & 77.33 & 73.59 & 71.72 \\
    5, 40  & \includegraphics[width=1cm, frame]{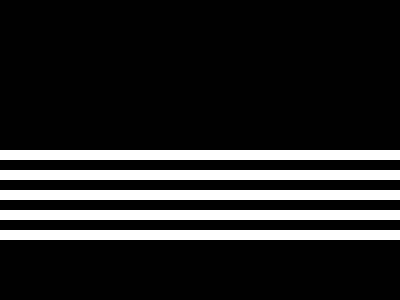} & 5123       & 90     & 77.28   & 77.2  & 78.82 & 75.72 \\
    2, 40  & \includegraphics[width=1cm, frame]{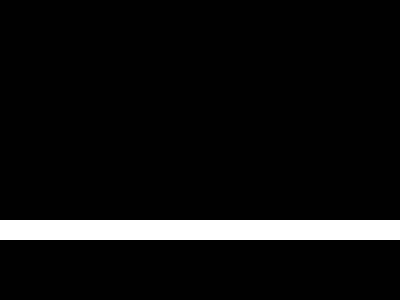} & 5123       & 110    & 67.54   & 66.66 & 65.88 & 70.2  \\
    8, 14  & \includegraphics[width=1cm, frame]{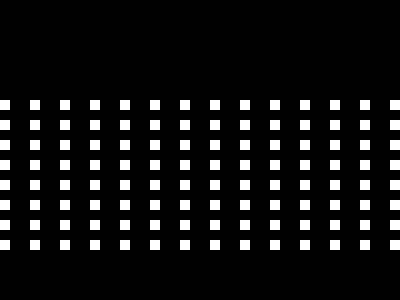} & 4867       & 58     & 76.21   & 78.26 & 67.05 & 83.72 \\
    3, 6   & \includegraphics[width=1cm, frame]{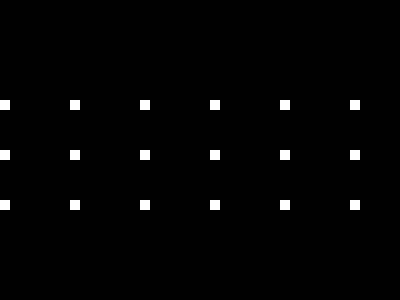} & 4099       & 48     & 68.53   & 68.26 & 68.23 & 69.1
  \end{tabular}
  \caption{Comparison of different input image masks on the base layout of Table \ref{tab:general_architecture}}
  \label{tab:different_masks}
\end{table}

The final network architecture is visualized in Table \ref{tab:final_architecture}.

\begin{table}
  \begin{tabular}{l|l|l|l}
    Type  & Shape        & Parameters & Description                             \\
    \hline
    Input & (15, 20, 2)  & 0          & \begin{tabular}[c]{@{}l@{}}Optical flow input layer \\ downscaled x, y, (dx, dy)\end{tabular} \\
    Conv  & (15, 20, 32) & 608        & \begin{tabular}[c]{@{}l@{}}2D Convolutions, kernel (3, 3) \\ 32 filters \end{tabular}         \\
    Pool  & (8, 10, 32) & 0          & Max-Pooling, kernel (2, 2)              \\
    Conv  & (8, 10, 8)  & 2312       & \begin{tabular}[c]{@{}l@{}}2D Convolutions, kernel (3, 3) \\ 8 filters \end{tabular} \\
    Conv  & (8, 10, 8)  & 584        & \begin{tabular}[c]{@{}l@{}}2D Convolutions, kernel (3, 3) \\ 8 filters \end{tabular}          \\
    Pool  & (4, 5, 8)    & 0          & Max-Pooling, kernel (2, 2)              \\
    Dense & 32           & 5152       &                                         \\
    Dense & 16           & 528        &                                         \\
    Dense & 3            & 51         &
  \end{tabular}
  \caption{Final architecture (9235 Parameters)}
  \label{tab:final_architecture}
\end{table}

\subsection{Deployment}
\label{sec:deployment}
After the network was designed and trained, it has to be deployed to the robot. The network can either be executed on the robot or on a remote computer with the robot transferring image data and receiving control commands instead of determining them on its own. Both options were evaluated.

For the deployment on the car, a python implementation was pursued. The first approach consisted of linear workflow: Each time a new frame was parsed, the inference in Keras was started and afterwards the processed results sent to the car. Unfortunately, the inference in Keras could only be processed with about 10 to 15 FPS. This blocked the parsing of the frames and resulted in a delay that increased with time. Therefore, the solution was implemented with three threads: The first thread continuously parses the frames and preprocesses them, the second thread runs the inference on the Keras module and the third thread manages the communication to the Arduino and transfers the speed as well as steer commands. The three threads communicate with Queues and Mutexes. This way, the inference thread could take a variable amount of time without blocking the parsing of the frames, effectively allowing to skip frames according to the current inference processing frame rate.

Unfortunately, the resulting frame rate of only up to 15 FPS is too slow for the speed the car was supposed to drive and with which the data was labeled with. During the labeling of the training data it was assumed that the data can be processed with 30 FPS. A slower processing rates results in slower reactions. Effectively, the car was capable of detecting obstacles, but it always happened too late (i.e. the car did steer, but only shortly after it crashed into an obstacle). Surprisingly, even an extremely small network with only about a 1000 parameters did not perform better. It seems like there is a large overhead on each prediction call.

It was attempted to cross compile an inference framework in C++ \cite{frugallydeep}, but this efforts were not completed since compiling on the Raspberry Pi was not possible due to the complexity of the framework and memory constraints and cross compiling did not lead to success either.

\begin{figure}[ht]
  \centering
  \begin{tikzpicture}[node distance=1.5cm,auto,>=latex']
      \node[block] (raspivid) {\texttt{raspivid}};
      \node[block] (proc_flow) [below of=raspivid] {\texttt{parse\_vectorfield}};
      \node[block] (inference) [below of=proc_flow] {\texttt{inference}};
      \node[block] (car) [below of=inference] {\texttt{car}};
      \path[-] (raspivid) edge node {\texttt{stdin} or UDP} (proc_flow);
      \path[-] (proc_flow) edge node {Queue/Mutex} (inference);
      \path[-] (inference) edge node {Queue/Mutex or UDP} (car);
  \end{tikzpicture}
  \caption{Visualization of the application software architecture}
  \label{fig:deployment_architecture}
\end{figure}
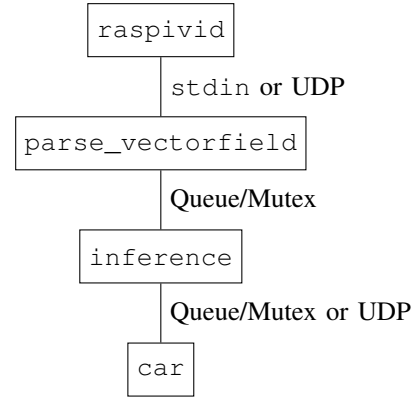

Therefore, the second approach was implemented in which the car transfers the preprocessed optical flow to a more powerful remote computer that is connected via a WiFi network. This computer is capable of processing the optical flow with up to 600 FPS. The camera interface on the Raspberry Pi directly allows to transfer the frame data remotely via a UDP or a TCP connection. On the end of the computer, the data stream behaves exactly as if it was processed on the Raspberry Pi. The steering commands are then processed on the computer and transferred back to the Raspberry Pi via a different UDP connection. The whole process is visualized in Fig. \ref{fig:deployment_architecture}.

This way, the car was generally capable of avoiding obstacles. Unfortunately, the WiFi module of the Raspberry Pi is extremely weak and an external antenna was not available, resulting in a large package loss. This resulted in the application working periodically in a period of about 5 seconds. For the purposes of this work as a proof of concept, this suffices.

\section{Results and Discussion}
Figure \ref{fig:eval_regression} shows the loss and a histogram of the labels vs. the predictions the in the test set for balanced and unbalanced data. Ideally, the histogram would be a perfectly straight line with a slope of 1 and a y-intercept of 0 (meaning that the predictions exactly match the labels). After only 22 epochs, the loss has decreased to 0.0116 for the unbalanced data and after 54 epochs to 0.0096 for the balanced data. The two histograms show an almost equal distribution, but the best fit linear curve has a slightly larger slope for the unbalanced data. The balancing does affect the performance positively because the loss is smaller for the balanced data as well and the learning takes a longer time.

The large variance especially around the label 0.5 in both histograms can be explained due to the inaccurate automatic labeling of the data, specifically in the transition between overriding the desired steering and passing it through. Furthermore, the human operator likely is not completely consistent with avoiding the obstacle, but the neural network learned a better point of deviating from the desired steering, resulting in a higher deviation from the labeled data.

Even though the regression approach is the favorable solution, the classification approach performs better. Most likely, the network has problems with properly learning the correlation between desired and corrected steering signal in the regression approach and requires a lot more training data to perform better, but likely definitely would if this training data was available. The classification approach is way broader than the regression. Only one of three classes have to be predicted without having to consider a desired steering direction at all (this only happens in the post processing). The classification also likely handles wrongly labeled data better than the regression approach, since a slightly wrong label has no effect at all to a classification in contrast to a regression.

Table \ref{tab:classification_accuracy} shows the accuracy for the three classes compared in balanced and unbalanced as well as automatically and manually labeled data. The problem with all combinations except the manual/unbalanced option is that the robot, when test driving in a real world environment, very often detected false positives, i.e. avoiding obstacles even though there are no obstacles. Obstacles were usually recognized, but as soon as the car started steering, it never stopped stering in the chosen direction. This was especially obvious in the automatically labeled training data. This is likely caused by wrongly labeled data, especially for the cases that an obstacle was just avoided: when the data was recorded, the operator usually kept the steering into the desired direction slightly longer than the obstacle was actually visible in the camera, resulting in the network to learn that if a turning direction is recognized by the camera to keep turning into this direction. The manual/unbalanced option has the highest accuracy for the classification of no obstacle and learned to not behave like this. Since the training data is as described, the accuracy for the left and right label is smaller, since this configuration learned not to keep turning slightly longer than the obstacle is actually seen.

Table \ref{tab:effect_manipulate_layers} shows the effect of manipulating single layers in the network. All the manipulations do not seem to influence the performance of the network significantly. Likely, the network can be reduced significantly in size, but since a significantly smaller network does not appear to reduce the execution time significantly (refer Section \ref{sec:deployment}), an optimization was not performed at this point.

In contrast to that, varying the mask applied to the optical flow frame did influence the network performance significantly. Table \ref{tab:different_masks} shows the effect of different masks to the network. For the avoidance of obstacles on the ground, only a small narrow area in the middle of the image is relevant. Obviously, this would not be the case for other kind of vehicles. Additionally, the resolution in the vertical direction can be bisected. The best performing mask turned out to have a size of 15 by 20 vectors.

\section{Conclusions}
In this work, a machine learning approach to obstacle avoidance based on optical flow was introduced. It was evaluated on a for this purpose modified toy race car.

Even though the implemented solution generally works, there are different flaws. The biggest problem is the data acquisition. It is generally desirable to learn from a human driver/operator, but the operator should rather see exactly what the car sees. If it is not possible for the human operator to avoid collisions under these circumstances, the cameras would have to be modified, likely to have a larger field of view. With this method, a much larger data set should be recorded to have a large variance. In order to avoid a bias, it can also be considered to have two operators, one of which controls the desired steering direction and one that solely avoids collisions.

While an end-to-end solution seems desirable at first glance with the motivation of simply learning one single network that does all the work and that is capable of solving all subproblems independently of the environment at once, there are different downsides. The whole operation of the robot becomes a large black box, making it impossible to fine tune parts of the network. This makes testing and validation and debugging difficult and requires a vast amount of data. A better alternative is to break the problem down into multiple sub-problems that can easily be validated on their own. For this particular use case, it could be considered to detect obstacles first and based on that information have simple rules about how to avoid collisions or even learn them with a different network. This way, problems can be easily detected and fixed faster and the overall complexity decreases.

Another obvious disadvantage over the optical flow approach is that it only works on good lighting conditions. It also has yet to be evaluated how different driving speeds affect the performance of the network. It is likely that the performance will not be affected since a different speed should impact the whole image equally.

The limited computation power on the Raspberry Pi was a big problem. A more powerful computing unit or optimized software might lead to better results. It could be considered to use a small low power compute stick like the Intel Movidius stick \cite{ion_2015} to enhance the performance. A decentralized solution with a computer remote controlling the device is strongly dependent on a well performing wireless network, which is not given in search and rescue scenarios.

Furthermore, other representations of the network output could be evaluated. It could be considered to process the offset to a current steering signal. The network could infer the current steering signal from the current optical flow. Also, the regression and the classification could be combined by processing a linear steering signal together with a likelihood of currently seeing an obstacle. Based on this likelihood, the steering prediction could be used or the desired steering. Lastly, the amount of classes that the classification approach currently predicts could be extended. This would make the labeling more complicated.

Generally the implemented solution works and demonstrates that it is possible to avoid obstacles solely with obstacle flow. It is to be expected that applying this concept to flying robots would lead to good results, but the network architecture and the procedure to obtain the training data would have to be reworked.

\bibliographystyle{IEEEtran}
\bibliography{quote.bib}

\vspace{\fill}

\appendix
Explanation of the project directory:
\begin{enumerate}
    \item \texttt{car}: Everything related to the software running on the car
    \begin{enumerate}
        \item \texttt{Bridge\_Car\_Raspi}: The software running on the Arduino that controls the low level electronics
        \begin{enumerate}
            \item \texttt{Bridge\_Car\_Raspi.ino} Module that contains the main loop and all the general logic and controllers
            \item \texttt{comm.ino}: Simple implementation of communication protocol
        \end{enumerate}
        \item \texttt{cpp}: The modules that were implemented in C++ for potentially running on the car
        \begin{enumerate}
            \item \texttt{CarControl.[h/cpp]}: Implementation of communication protocol to the Arduino
            \item \texttt{inference.cpp}: Implementation of C++ inference with frugally
            \item \texttt{record\_data.cpp}: Tool for recording training data on the Raspberry Pi. Particularly useful when not only motion vectors, but also raw video should be recorded (in that case the Python implementation is extremely slow)
        \end{enumerate}
        \item \texttt{python}: The modules that were implemented in Python for running on the car
        \begin{enumerate}
            \item \texttt{carcontrol.py}: Implementation of communication protocol to the Arduino
            \item \texttt{inference.py}: Implementation of Python inference with Keras
            \item \texttt{record\_data.py}: Tool for recording training data on the Raspberry Pi. If raw video data must be recorded additionally, the C++ implementation should be used!
            \item \texttt{remote\_car.py}: Simple bridge for remote controlling the car with the remote control
            \item \texttt{remote\_interface.py}: Script to be executed if the inference should run on another host via the local network
        \end{enumerate}
    \end{enumerate}
    \item \texttt{demo.mp4}: A small video demonstrating the general functionality of the system
    \item \texttt{doc}: This document as \LaTeX \:source
    \item \texttt{host}: Everything related to the software running on a host computer
    \begin{enumerate}
        \item \texttt{remote\_inference.py}: Inference (Keras) to be run on a remote host via the local network
        \item \texttt{obstacleavoidance.ipynp} Jupyter notebook used for training
        \item \texttt{data.hdf} HDF Archive containing all training data
        \item \texttt{visualize\_hdf.py} Tool for visualizing and labeling HDF data
        \item \texttt{convert\_raw\_hdf.py} Convert data recorded with record\_data.cpp into HDF archive
        \item \texttt{convert\_raspi\_h264\_mp4.sh} Convert raspivid .h264 file into readable MP4
    \end{enumerate}
\end{enumerate}
\end{document}

%% file: img/plots/regression_training_loss_unbalanced.tex
\begin{tikzpicture}

\definecolor{color0}{rgb}{0.12156862745098,0.466666666666667,0.705882352941177}
\definecolor{color1}{rgb}{1,0.498039215686275,0.0549019607843137}

\begin{axis}[
legend cell align={left},
legend entries={{Train},{Test}},
legend style={draw=white!80.0!black},
tick align=outside,
tick pos=left,
width=\figurewidth,
x grid style={white!69.01960784313725!black},
xmin=-1.05, xmax=22.05,
y grid style={white!69.01960784313725!black},
ymin=0.0103737880970311, ymax=0.0362139542080583
]
\addlegendimage{no markers, color0}
\addlegendimage{no markers, color1}
\addplot [semithick, color0]
table [row sep=\\]{%
0	0.0350394012030116 \\
1	0.0169613948823523 \\
2	0.0161990205344239 \\
3	0.0153472348679924 \\
4	0.0149656387007994 \\
5	0.0145208881339575 \\
6	0.0140252413197236 \\
7	0.0137188894236982 \\
8	0.0135148749471274 \\
9	0.0132273617826199 \\
10	0.0129891221225796 \\
11	0.0127585385863133 \\
12	0.0125298170681707 \\
13	0.0124495248862995 \\
14	0.0122291237704956 \\
15	0.0121222691559614 \\
16	0.0119709153907006 \\
17	0.0119479031199791 \\
18	0.0117634782966005 \\
19	0.0117284800021675 \\
20	0.0117065471888984 \\
21	0.0115483411020778 \\
};
\addplot [semithick, color1]
table [row sep=\\]{%
0	0.016928570472834 \\
1	0.0152652689034475 \\
2	0.0150679371022313 \\
3	0.0136283252624794 \\
4	0.0139716064094733 \\
5	0.0130986449672859 \\
6	0.0127107407470548 \\
7	0.0127189297444069 \\
8	0.0124990445398201 \\
9	0.0122819308842692 \\
10	0.012603614826777 \\
11	0.0125144750354024 \\
12	0.0128767746374137 \\
13	0.0118844646235038 \\
14	0.0117812874761232 \\
15	0.0115959278176296 \\
16	0.0115912685843461 \\
17	0.0120006317992857 \\
18	0.0116739527224176 \\
19	0.0118111187545843 \\
20	0.0121424441712397 \\
21	0.0116074478368948 \\
};
\end{axis}

\end{tikzpicture}

%% file: img/plots/regression_evaluation_unbalanced.tex
\begin{tikzpicture}

\begin{axis}[
tick align=outside,
tick pos=left,
width=\figurewidth,
x grid style={white!69.01960784313725!black},
xmin=0.00784313725490196, xmax=1,
y grid style={white!69.01960784313725!black},
ymin=0, ymax=1
]
\addplot graphics [includegraphics cmd=\pgfimage,xmin=0.00784313725490196, xmax=1, ymin=0, ymax=1] {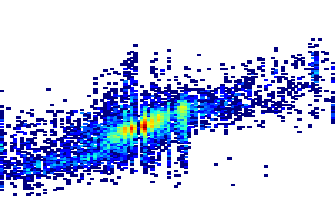};
\addplot [line width=1.0pt, red, forget plot]
table [row sep=\\]{%
0	0.234838411016513 \\
1	0.660755792497908 \\
};
\end{axis}

\end{tikzpicture}

%% file: img/plots/regression_training_loss_balanced.tex
\begin{tikzpicture}

\definecolor{color0}{rgb}{0.12156862745098,0.466666666666667,0.705882352941177}
\definecolor{color1}{rgb}{1,0.498039215686275,0.0549019607843137}

\begin{axis}[
legend cell align={left},
legend entries={{Train},{Test}},
legend style={draw=white!80.0!black},
tick align=outside,
tick pos=left,
width=\figurewidth,
x grid style={white!69.01960784313725!black},
xmin=-2.65, xmax=55.65,
y grid style={white!69.01960784313725!black},
ymin=0.00827886406758828, ymax=0.0305053726518774
]
\addlegendimage{no markers, color0}
\addlegendimage{no markers, color1}
\addplot [semithick, color0]
table [row sep=\\]{%
0	0.029495076807137 \\
1	0.0192320254679575 \\
2	0.0168998809930391 \\
3	0.0158299437494582 \\
4	0.0149920925395591 \\
5	0.0145208516248931 \\
6	0.0139929588403557 \\
7	0.0136557818172954 \\
8	0.0133533573322712 \\
9	0.0131735026124234 \\
10	0.0130455044864146 \\
11	0.0127110894970299 \\
12	0.0125792403471683 \\
13	0.0123571422353189 \\
14	0.0123470059175675 \\
15	0.0122616813727948 \\
16	0.0119994788642257 \\
17	0.0119119794634188 \\
18	0.0116984118256515 \\
19	0.0116592474392484 \\
20	0.0114715362557896 \\
21	0.0113965127002023 \\
22	0.0113439919567725 \\
23	0.0111108603469114 \\
24	0.0110036680598214 \\
25	0.0109487634098437 \\
26	0.0108762174424266 \\
27	0.0107865004383899 \\
28	0.0107114598151403 \\
29	0.0107117901737779 \\
30	0.0105201506262659 \\
31	0.0105881268371123 \\
32	0.0104912469808875 \\
33	0.0104228947948649 \\
34	0.010419956951411 \\
35	0.010220862222756 \\
36	0.0103375504204614 \\
37	0.0101868053116454 \\
38	0.0101502129180775 \\
39	0.010226350068515 \\
40	0.00989754236206454 \\
41	0.00996732972817117 \\
42	0.00981898120937931 \\
43	0.00970017319803325 \\
44	0.00963092625513732 \\
45	0.00981492671015843 \\
46	0.00961317632160159 \\
47	0.00954977043525579 \\
48	0.00960317223502411 \\
49	0.00958354044257492 \\
50	0.00937199654975114 \\
51	0.00928915991232869 \\
52	0.00944312035163181 \\
53	0.00943794161479482 \\
};
\addplot [semithick, color1]
table [row sep=\\]{%
0	0.0222095211617663 \\
1	0.0241883651408739 \\
2	0.0177469390456553 \\
3	0.0177876462385436 \\
4	0.0137531860081778 \\
5	0.012584119364191 \\
6	0.0121029656176583 \\
7	0.011611302511928 \\
8	0.0117190385737796 \\
9	0.0115183275341686 \\
10	0.0119644322555313 \\
11	0.0114495855916116 \\
12	0.0114866395274521 \\
13	0.0114223683664957 \\
14	0.0112608086872169 \\
15	0.0112600384582661 \\
16	0.0112830017458775 \\
17	0.0112352460168329 \\
18	0.0111718203088045 \\
19	0.0109173828225971 \\
20	0.0109824009486731 \\
21	0.0110016322278528 \\
22	0.010783332233865 \\
23	0.0108902348887135 \\
24	0.0107398827381595 \\
25	0.0110847330623427 \\
26	0.0106081138393721 \\
27	0.0109635043067611 \\
28	0.0108245158746873 \\
29	0.010673439313398 \\
30	0.0105317165077364 \\
31	0.0105258937578506 \\
32	0.0108058384816149 \\
33	0.0106635354528684 \\
34	0.0103801563310768 \\
35	0.0106031017880415 \\
36	0.0102626556599596 \\
37	0.0103090279762393 \\
38	0.0102529875969418 \\
39	0.0101023317212173 \\
40	0.0100040771917765 \\
41	0.0101912735672087 \\
42	0.0103220182980756 \\
43	0.00999481619920836 \\
44	0.0103070764319187 \\
45	0.0103928372938061 \\
46	0.0101659955634909 \\
47	0.0099053793551695 \\
48	0.00966156861089108 \\
49	0.0097432995236076 \\
50	0.00998151416983317 \\
51	0.0099600381292252 \\
52	0.00969074407843439 \\
53	0.00967736666189728 \\
};
\end{axis}

\end{tikzpicture}

%% file: img/plots/regression_evaluation_balanced.tex
\begin{tikzpicture}

\begin{axis}[
tick align=outside,
tick pos=left,
width=\figurewidth,
x grid style={white!69.01960784313725!black},
xmin=0.00784313725490196, xmax=1,
y grid style={white!69.01960784313725!black},
ymin=0, ymax=1
]
\addplot graphics [includegraphics cmd=\pgfimage,xmin=0.00784313725490196, xmax=1, ymin=0, ymax=1] {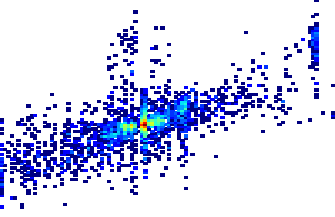};
\addplot [line width=1.0pt, red, forget plot]
table [row sep=\\]{%
0	0.210690832155833 \\
1	0.720248492698338 \\
};
\end{axis}

\end{tikzpicture}

%% file: ms.bbl
\begin{thebibliography}{10}
\providecommand{\url}[1]{#1}
\csname url@samestyle\endcsname
\providecommand{\newblock}{\relax}
\providecommand{\bibinfo}[2]{#2}
\providecommand{\BIBentrySTDinterwordspacing}{\spaceskip=0pt\relax}
\providecommand{\BIBentryALTinterwordstretchfactor}{4}
\providecommand{\BIBentryALTinterwordspacing}{\spaceskip=\fontdimen2\font plus
\BIBentryALTinterwordstretchfactor\fontdimen3\font minus
  \fontdimen4\font\relax}
\providecommand{\BIBforeignlanguage}[2]{{%
\expandafter\ifx\csname l@#1\endcsname\relax
\typeout{** WARNING: IEEEtran.bst: No hyphenation pattern has been}%
\typeout{** loaded for the language `#1'. Using the pattern for}%
\typeout{** the default language instead.}%
\else
\language=\csname l@#1\endcsname
\fi
#2}}
\providecommand{\BIBdecl}{\relax}
\BIBdecl

\bibitem{dak_2016}
R.~Dakin, T.~Fellows, and D.~L.~Altshuler, ``Visual guidance of forward flight
  in hummingbirds reveals control based on image features instead of pattern
  velocity,'' \emph{Proceedings of the National Academy of Sciences}, vol. 113,
  pp. 8849--8854, 07 2016.

\bibitem{ber_1981}
\BIBentryALTinterwordspacing
B.~K.P.~Horn and B.~G. Schunck, ``Determining optical flow,'' \emph{Artificial
  Intelligence}, vol.~17, no.~1, pp. 185--203, 1981. [Online]. Available:
  \url{http://www.sciencedirect.com/science/article/pii/0004370281900242}
\BIBentrySTDinterwordspacing

\bibitem{lon_1980}
H.~C~Longuet-Higgins and K.~Prazdny, ``The interpretation of a moving retinal
  image,'' \emph{Proceedings of the Royal Society of London. Series B,
  Containing papers of a Biological character. Royal Society (Great Britain)},
  vol. 208, pp. 385--97, 08 1980.

\bibitem{bar_2002}
G.~L.~Barrows, J.~S.~Chahl, and M.~V~Srinivasan, ``Biomimetic visual sensing
  and flight control,'' \emph{Proc. Bristol UAV Conf}, pp. 159--168, 01 2002.

\bibitem{gre_2003}
W.~E. Green, P.~Y. Oh, K.~Sevcik, and G.~Barrows, ``Autonomous landing for
  indoor flying robots using optic flow,'' \emph{ASME International Mechanical
  Engineering Congress and Exposition}, no. 37130, pp. 1347--1352, 12 2003.

\bibitem{bey_2009}
A.~Beyeler, J.-C. Zufferey, and D.~Floreano, ``Vision-based control of
  near-obstacle flight,'' \emph{Autonomous Robots}, vol.~27, no.~3, pp.
  201--219, 08 2009.

\bibitem{mcg_2017}
K.~McGuire, G.~de~Croon, C.~De~Wagter, K.~Tuyls, and H.~Kappen, ``Efficient
  optical flow and stereo vision for velocity estimation and obstacle avoidance
  on an autonomous pocket drone,'' \emph{IEEE Robotics and Automation Letters},
  vol.~2, no.~2, pp. 1070--1076, 04 2017.

\bibitem{san_2018}
N.~J. Sanket, C.~D. Singh, K.~Ganguly, C.~Fermüller, and Y.~Aloimonos,
  ``Gapflyt: Active vision based minimalist structure-less gap detection for
  quadrotor flight,'' \emph{IEEE Robotics and Automation Letters}, vol.~3,
  no.~4, pp. 2799--2806, 10 2018.

\bibitem{mic_2005}
J.~Michels, A.~Saxena, and A.~Y.~Ng, ``High speed obstacle avoidance using
  monocular vision and reinforcement learning,'' in \emph{ICML 2005 -
  Proceedings of the 22nd International Conference on Machine Learning}, 01
  2005, pp. 593--600.

\bibitem{tai_2016}
L.~Tai, S.~Li, and M.~Liu, ``A deep-network solution towards model-less
  obstacle avoidance,'' in \emph{2016 IEEE/RSJ International Conference on
  Intelligent Robots and Systems (IROS)}, 10 2016, pp. 2759--2764.

\bibitem{ost_2004}
J.~Ostermann, J.~Bormans, P.~List, D.~Marpe, M.~Narroschke, F.~Pereira,
  T.~Stockhammer, and T.~Wedi, ``Video coding with h.264/avc: tools,
  performance, and complexity,'' \emph{Circuits and Systems Magazine, IEEE},
  vol.~4, pp. 7--28, 09 2004.

\bibitem{hol_2014}
\BIBentryALTinterwordspacing
G.~Hollingworth, ``Vectors from coarse motion estimation,'' 2014, retrieved
  02/10/19 5:27 pm. [Online]. Available:
  \url{https://www.raspberrypi.org/blog/vectors-from-coarse-motion-estimation/}
\BIBentrySTDinterwordspacing

\bibitem{hei_2017}
\BIBentryALTinterwordspacing
A.~Heinrich, ``An optical flow odometry sensor based on the raspberry pi
  computer,'' Master's thesis, Czech Technical University in Prague, 01 2017.
  [Online]. Available:
  \url{https://dspace.cvut.cz/bitstream/handle/10467/67320/F3-DP-2017-Heinrich-Adam-DP_Adam_Heinrich_2017_01.pdf}
\BIBentrySTDinterwordspacing

\bibitem{ras_2019}
\BIBentryALTinterwordspacing
R.~P. Foundation, ``Raspberry pi zero,'' 2019, retrieved 02/18/19 3:24 pm.
  [Online]. Available:
  \url{https://www.raspberrypi.org/products/raspberry-pi-zero/}
\BIBentrySTDinterwordspacing

\bibitem{sie_2004}
R.~Siegwart and I.~R. Nourbakhsh, \emph{Introduction to Autonomous Mobile
  Robots}.\hskip 1em plus 0.5em minus 0.4em\relax Scituate, MA, USA: Bradford
  Company, 2004.

\bibitem{apm_2019}
\BIBentryALTinterwordspacing
ArduPilot, ``Ardupilot,'' 2019, retrieved 02/18/19 4:17 pm. [Online].
  Available: \url{http://ardupilot.org/}
\BIBentrySTDinterwordspacing

\bibitem{dau_2012}
A.~D'Ausilio, ``Arduino: A low-cost multipurpose lab equipment,''
  \emph{Behavior Research Methods}, vol.~44, no.~2, pp. 305--313, 06 2012.

\bibitem{goo_2016}
I.~Goodfellow, Y.~Bengio, and A.~Courville, \emph{Deep Learning}.\hskip 1em
  plus 0.5em minus 0.4em\relax MIT Press, 2016,
  \url{http://www.deeplearningbook.org}.

\bibitem{sim_2014}
\BIBentryALTinterwordspacing
K.~Simonyan and A.~Zisserman, ``Very deep convolutional networks for
  large-scale image recognition,'' \emph{CoRR}, vol. abs/1409.1556, 2014.
  [Online]. Available: \url{http://arxiv.org/abs/1409.1556}
\BIBentrySTDinterwordspacing

\bibitem{yan_2005}
\BIBentryALTinterwordspacing
Y.~LeCun, U.~Muller, J.~Ben, E.~Cosatto, and B.~Flepp, ``Off-road obstacle
  avoidance through end-to-end learning,'' in \emph{Proceedings of the 18th
  International Conference on Neural Information Processing Systems}, ser.
  NIPS'05.\hskip 1em plus 0.5em minus 0.4em\relax Cambridge, MA, USA: MIT
  Press, 2005, pp. 739--746. [Online]. Available:
  \url{http://dl.acm.org/citation.cfm?id=2976248.2976341}
\BIBentrySTDinterwordspacing

\bibitem{bat_2004}
G.~E. A. P.~A. Batista, R.~C. Prati, and M.~C. Monard, ``A study of the
  behavior of several methods for balancing machine learning training data,''
  \emph{SIGKDD Explor. Newsl.}, vol.~6, no.~1, pp. 20--29, 06 2004.

\bibitem{frugallydeep}
\BIBentryALTinterwordspacing
T.~Hermann, ``frugally-deep: Header-only library for using keras models in
  c++,'' 2018, retrieved 04/19/19 20:24 pm. [Online]. Available:
  \url{https://github.com/Dobiasd/frugally-deep}
\BIBentrySTDinterwordspacing

\bibitem{ion_2015}
M.~H. Ionica and D.~Gregg, ``The movidius myriad architecture's potential for
  scientific computing,'' \emph{IEEE Micro}, vol.~35, no.~1, pp. 6--14, Jan
  2015.

\end{thebibliography}
